\documentclass[10pt,twocolumn,letterpaper]{article}

\usepackage[pagenumbers]{cvpr} 

\definecolor{cvprblue}{rgb}{0.21,0.49,0.74}
\usepackage[pagebackref,breaklinks,colorlinks,allcolors=cvprblue]{hyperref}

\usepackage{multicol}

\usepackage{times}
\usepackage{epsfig}
\usepackage{graphicx}
\usepackage{amsmath}
\usepackage{amssymb}
\usepackage{comment}
\usepackage{float}
\usepackage{booktabs}
\usepackage{array}
\usepackage{multirow}
\usepackage[table]{xcolor}
\usepackage{caption}
\usepackage{subcaption}
\usepackage{framed}
\usepackage[export]{adjustbox}
\usepackage{varwidth}
\usepackage{algorithm}
\usepackage{algpseudocode}
\usepackage{makecell}
\usepackage{placeins}
\usepackage{longtable}
\usepackage{graphicx}
\usepackage{multirow}
\usepackage{rotating}

\def\paperID{*****} 
\def\confName{CVPR}
\def\confYear{2026}

\title{Fusion-CAM: Integrating Gradient and Region-Based Class Activation Maps for Robust Visual Explanations}

\author{Hajar Dekdegue\\
\small \textit{IRIT, UMR5505 CNRS}\\ 
\small \textit{Université de Toulouse}\\
{\small hajar.dekdegue@irit.fr}
\and
Moncef Garouani\\
\small \textit{IRIT, UMR5505 CNRS}\\ 
\small \textit{Université Toulouse Capitole}\\
{\small moncef.garouani@irit.fr}
\and
Josiane Mothe\\
\small \textit{IRIT, UMR5505 CNRS}\\ 
\small \textit{Université de Toulouse}\\
{\small josiane.mothe@irit.fr}
\and
Jordan Bernigaud \\
\small \textit{INRAE, Centre Occitanie-Toulouse}\\ \small\textit{ Unité Expérimentale APC}\\
{\small jordan.bernigaud-samatan@inrae.fr}
}

\begin{document}

\maketitle

\begin{abstract}

Interpreting the decision-making process of deep convolutional neural networks remains a central challenge in achieving trustworthy and transparent artificial intelligence. Explainable AI (XAI) techniques, particularly Class Activation Map (CAM) methods, are widely adopted to visualize the input regions influencing model predictions. Gradient-based approaches (e.g. Grad-CAM) provide highly discriminative, fine-grained details by computing gradients of class activations but often yield noisy and incomplete maps that emphasize only the most salient regions rather than the complete objects. Region-based approaches (e.g. Score-CAM) aggregate information over larger areas, capturing broader object coverage at the cost of over-smoothing and reduced sensitivity to subtle features.
We introduce Fusion-CAM, a novel framework that bridges this explanatory gap by unifying both paradigms through a dedicated fusion mechanism to produce robust and highly discriminative visual explanations. Our method first denoises gradient-based maps, yielding cleaner and more focused activations. It then combines the refined gradient map with region-based maps using contribution weights to enhance class coverage. Finally, we propose an adaptive similarity-based pixel-level fusion that evaluates the agreement between both paradigms and dynamically adjusts the fusion strength. This adaptive mechanism reinforces consistent activations while softly blending conflicting regions, resulting in richer, context-aware, and input-adaptive visual explanations.
Extensive experiments on standard benchmarks show that Fusion-CAM consistently outperforms existing CAM variants in both qualitative visualization and quantitative evaluation, providing a robust and flexible tool for interpreting deep neural networks.

\end{abstract}
\vspace{-0.2cm}
\section{Introduction}
Despite their remarkable performance in computer vision tasks, deep convolutional neural networks remain fundamentally opaque in their decision-making processes~\cite{zhao2024review,blackBox,TwoDimensions,OpacityDNNs, Adadi2018}. This opacity becomes particularly problematic in safety-critical applications such as medical diagnosis, autonomous driving, and security systems, where understanding why a model produces a specific prediction is as important as the prediction itself~\cite{MedicalBlackBox}. Explainable Artificial Intelligence (XAI) addresses this challenge by developing methods to make model behaviors transparent and interpretable~\cite{Samek2017XAI,Mershaa2023XAI,xai_medical_review2024}.

Among XAI approaches, post-hoc methods have become a dominant line of research for interpreting trained models without modifying their architecture~\cite{PostHocXAI,LIME,SHAP}. Within this paradigm, saliency-based approaches generate visual explanations as heatmaps that highlight regions of an input image most responsible to the model’s decision~\cite{Grad-CAM,Grad-CAM++, XGrad-CAM, RISE, Ablation-CAM, Score-CAM, Group-CAM}. Class Activation Maps (CAM)~\cite{CAM} pioneered this direction by producing class-specific localization maps, laying the groundwork for subsequent visual explanation techniques. However, CAM's architectural dependence—requiring a global average pooling (GAP) layer before the final classification, limits its applicability  to specific network designs such as VGG~\cite{VGG16} and AlexNet~\cite{AlexNet}, that rely on fully connected layers. 

To overcome these limitations, two distinct families of methods have emerged. Gradient-based methods such as Grad-CAM \cite{Grad-CAM} operates at the pixel level and use backpropagation to measure the contribution of each pixel to the predicted class, producing highly class-discriminative maps. Yet, their reliance on gradient signals often results in noisy activations that fail to capture the full extent of  target objects, especially in multi-instance scenarios~\cite{Grad-CAM++}. 
On the other hand, gradient-free (or region-based) methods like Score-CAM~\cite{Score-CAM} generate activation maps 
by masking regions of the input image and measuring their impact on class scores. While this approach produces spatially broader coverage, it  may overlook fine-grained, class-specific details~\cite{SS-CAM}.

These complementary characteristics (gradient-based methods excelling at pixel-level precision and region-based methods capturing broader semantic context) 
 motivate their integration. 
 Our Fusion-CAM  achieves this through a novel adaptive fusion mechanism that combines the complementary strengths of the two map types, the process begins by denoising gradient-based map to eliminate  background noise and establish a pixel level precision baseline. This map is then linearly aggregated with the broader contextual coverage of region-based map, using their contribution weights to the model's decision. In the final step Fusion-CAM dynamically adjusting the fusion rule at a pixel level based on local map similarity; pixels with high similarity adopt maximum activation, while conflicting regions are blended proportionally, producing precise and input-adaptive explanations.

\begin{figure*}[!]
    \centering
            \centering
            \includegraphics[height=0.45\textwidth]{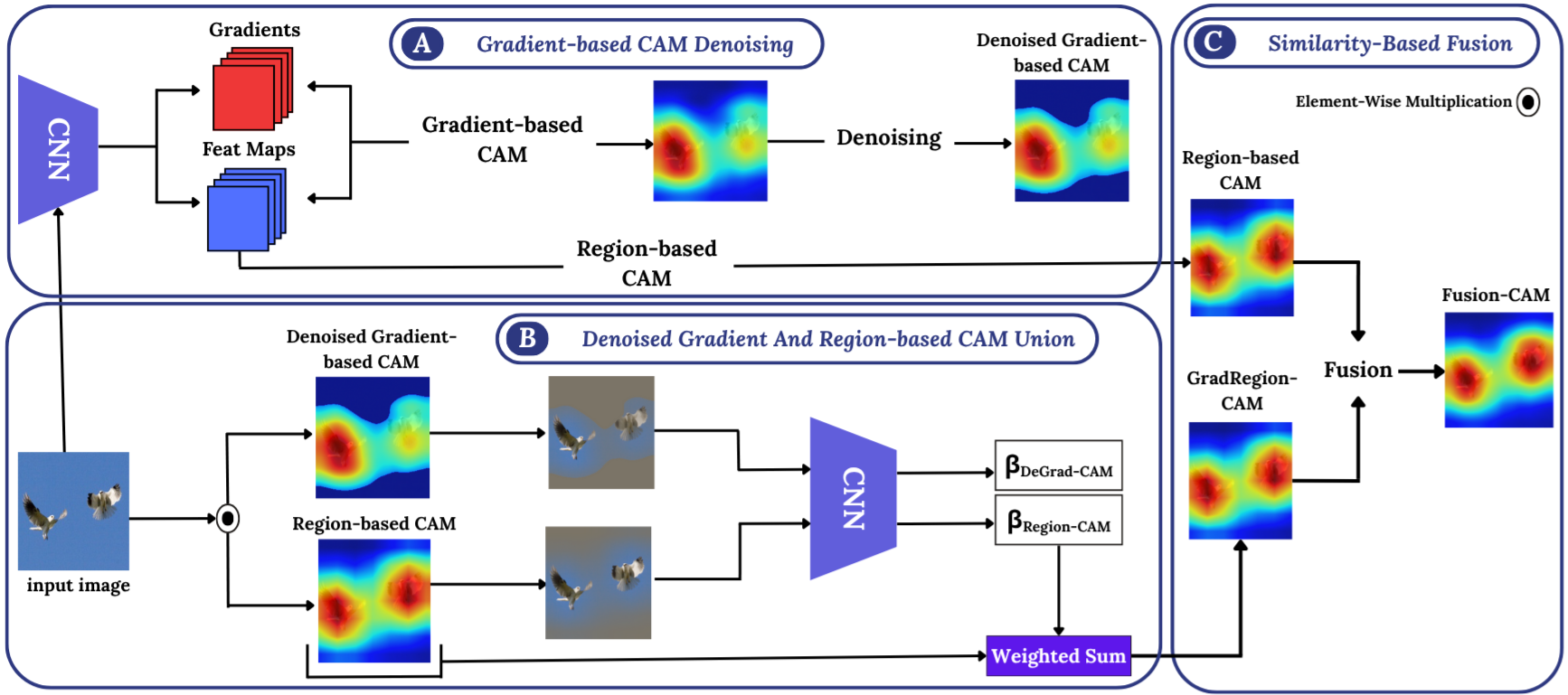}
            \vspace{-9pt}
            \caption{Overview of  Fusion-CAM.} 
            \vspace{-9pt}\label{fig:workflow}
\end{figure*}
 We validate our approach through extensive experiments 
 on standard benchmarks including ImageNet~\cite{imagenet}, PASCAL VOC~\cite{VOC2007}, and domain-specific datasets for plant disease detection. While we use Grad-CAM~\cite{Grad-CAM} and Score-CAM~\cite{Score-CAM} as representative instantiations of gradient-based and region-based methods, Fusion-CAM presents a generic framework, establishing it as a flexible fusion framework rather than a method-specific combination.

The main contributions of this work are as follows:
\begin{itemize}
\item \textbf{Fusion-CAM:} a post-hoc saliency method unifying gradient-based and gradient-free CAMs via multi-stage fusion (denoising, confidence-weighted aggregation, similarity-aware pixel blending).
\item \textbf{Stronger results:} consistent visual and quantitative gains across datasets (ImageNet/VOC and plant-disease sets) and architectures (VGG16, ResNet50, MobileNet), outperforming prior CAMs on Average Drop/Increase and Deletion–Insertion AUC.
\item \textbf{Ablations \& robustness:} each fusion stage contributes; explanations are more robust to noise and class confusion than individual CAM variants.
\end{itemize}

\section{Related Work}
\label{sec:related work}
Class activation map (CAM) approaches can be broadly categorized into gradient-based~\cite{Grad-CAM,Grad-CAM++,XGrad-CAM}, region-based~\cite{RISE, Score-CAM,Ablation-CAM,Group-CAM}, and ensemble methods~\cite{GradPPScoreCAM,FD-CAM, Feature-CAM,Union-CAM}. 

\textbf{Gradient-Based Methods. } 
  Grad-CAM~\cite{Grad-CAM}  generates class-specific saliency maps by computing a gradient-weighted linear combination of feature maps at a given convolutional layer. The maps are often noisy and focus only on the most discriminative features.  Grad-CAM++~\cite{Grad-CAM++} extends Grad-CAM for multi-object localization by using positive partial derivatives of feature maps with respect to the class score as weights, generating more class-specific visual maps. XGrad-CAM~\cite{XGrad-CAM} improves Grad-CAM by weighting each feature map through the element-wise product of activations and their gradients, emphasizing regions that are both strongly activated and highly influential for the prediction. Overall, gradient-based methods operate at the pixel level, highlighting the most discriminative features for the model. However, they often produce noisy and incomplete localization maps due to gradient saturation problems, which limits their interpretability~\cite{Rakitianskaia2015,Buono2024}.

\textbf{Region-Based Methods. } RISE~\cite{RISE} generates thousands of random binary masks and applies each mask to the input image, creating a set of masked images. Each masked image is fed through the model to obtain a prediction score for the target class. The final saliency map is produced by linearly combining all masks, weighted by their corresponding prediction scores. Instead of generating random masks, Score-CAM  ~\cite{Score-CAM} uses up-sampled activation maps as masks and linearly combines them weighted by the model’s confidence score on the masked inputs. Group-CAM~\cite{Group-CAM} 
is much faster than Score-CAM, as it groups similar feature maps to mask the input instead of using individual ones, reducing the number of forward passes through the model. The grouped feature maps are then linearly combined with their corresponding confidence scores. Overall, region-based methods operate at the region level by evaluating the highlighted areas through masking or perturbation, providing more complete coverage of the target class object, but they tend to over-smooth, which can reduce boundary precision~\cite{SS-CAM}.
 
\textbf{Ensemble Methods}. 
Increment-CAM~\cite{Increment-CAM} refines Grad-CAM by applying Score-CAM to its output, producing a secondary heatmap that is then element-wise multiplied with the original Grad-CAM map to suppress false or irrelevant activations. While this operation effectively reduces noise, it can inadvertently attenuate or eliminate model-relevant regions, particularly when either heatmap exhibits near-zero activations, resulting in incomplete or fragmented localization of salient features.
FD-CAM~\cite{FD-CAM} ccombines the gradient-based weights of Grad-CAM with score-based weights computed via a grouped channel “on/off” switching mechanism similar to Ablation-CAM~\cite{Ablation-CAM} but applied to groups of channels rather than individual ones. However, the integration of these two weight types relies on heuristic design choices, empirically tuned for performance rather than derived from theoretical principles.
The very last and one of the most promising techniques in this line of research is Union-CAM~\cite{Union-CAM} 
the final heatmap is obtained by denoising the gradient-based map and linearly merging it with a region-based one. Its fusion mechanism, however, is fundamentally limited: by hardly choosing between the combined map and the region-based one solely based on confidence scores, it systematically discards valid activations. This can result in incomplete localization and consistent failure to capture certain relevant regions, particularly in complex or low-contrast scenarios.


Although ensemble methods generally improve visual interpretability over individual CAMs, they remain limited: relevant regions can be suppressed, merging schemes are often fixed, and maps are discarded solely based on confidence scores. Here, we hypothesize that the effectiveness of an ensemble CAM depends not only on combining multiple sources but also on explicitly modeling their agreement and disagreement at a fine-grained level. When two maps highlight overlapping regions, this consensus likely indicates reliable activations of target-related features; conversely, discrepancies may correspond to noise or complementary cues that should be retained. Building on this, Fusion-CAM performs pixel-level similarity-based fusion, reinforcing high-agreement pixels via the maximum and softly averaging low-agreement pixels. This adaptive strategy preserves complementary information while emphasizing consistent activations, yielding more informative explanations that faithfully reflect the model’s decisions.

\section{Fusion-CAM approach}
Fusion-CAM builds on a fundamental observation: gradient-based and gradient-free methods fail in opposite ways.
Gradient-based techniques achieve sharp class discrimination but are often noisy and spatially incomplete, while gradient-free methods produce coherent spatial coverage at the cost of class boundary precision. Rather than treating these as alternative approaches, we recognize them as complementary sources of evidence that, when properly reconciled, can overcome their individual limitations.

Our framework consists of three-step process, each designed to mitigate specific limitations of existing CAM methods while leveraging their complementary strengths. The first step, \textit{Gradient-Based CAM Denoising}, addresses the inherent noise in gradient-based maps, which primarily affects low-activation regions. By thresholding, background artifacts are removed, while discriminative activations are preserved, resulting in cleaner and more focused heatmaps. The second step, \textit{Combination of Denoised Gradient and Region-Based CAMs}, tackles spatial incompleteness by merging the denoised CAM with region-based activations, weighted by class confidence scores to prioritize reliable evidence. This stage enriches region-level maps with fine-grained details and restores the full spatial extent of the target object, as illustrated in Figure~\ref{fig:workflow}.

The core innovation resides in the final step, \textit{CAMs Fusion}, which goes beyond naive aggregation strategies such as element-wise multiplication~\cite{Increment-CAM}, heuristic combinations~\cite{FD-CAM}, or selecting a single map based on confidence scores~\cite{Union-CAM}. We hypothesize that meaningful agreement between gradient and region-based activation maps signals robust, class-relevant activations, whereas disagreement indicates boundaries or ambiguous features. Fusion-CAM operationalizes this intuition through a pixel-level similarity-based mechanism: at each pixel, local agreement between the two maps determines adaptive weighting. High-agreement regions adopt the maximum activation to reinforce consistent evidence, while low-agreement regions are softly averaged to retain nuanced information. The resulting fusion produces activation maps that are both precise and spatially coherent.

The following sections formalize each process-step and provide implementation details.

\textbf{Notation:} Let $f$ denote a convolutional neural network (CNN) that takes an RGB input image $I_0 \in \mathbb{R}^{3 \times H \times W}$, where $H$ and $W$ are the image height and width. The network produces a probability distribution $Y = f(I_0)$, with $Y^c$ representing the predicted probability for class $c$. Let $I_b \in \mathbb{R}^{3 \times H \times W}$ denote a black image of the same dimensions as $I_0$. For a given layer $l$ of the CNN, we denote its activations by $A^l$, with $A_k^l$ referring to the activation map of the $k$-th channel.

\label{sec:method}
\subsection{Gradient-based CAM Denoising}
\label{threshold}
Gradient-based CAM methods highlight image regions that contribute most to a target class by backpropagating gradients to intermediate feature maps. However, these gradients often degrade in two main ways. First, backpropagation amplifies high-frequency noise in deeper layers, causing false activations in background areas that are not related to the target class. Second, gradient saturation—arising from activation functions such as sigmoid or ReLU—can cause informative signals to vanish along certain network paths, resulting in incomplete coverage of the target object~\cite{Score-CAM}. These combined effects produce noisy and spatially fragmented activation maps that fail to fully capture class-discriminative regions.
To address this, we introduce a simple yet effective denoising strategy(see Part~A in Figure~\ref{fig:workflow}). Low-activation values, typically corresponding to irrelevant background areas, are filtered out by removing the bottom $\theta\%$ of pixel intensities in the gradient-based heatmap. Formally, for each spatial position $p$, the denoised map is defined as:
\vspace{-10pt}
\begin{equation}
L_{\text{DeGrad}}^c(p) = 
\begin{cases}
L_{\text{Grad}}^c(p), & \text{if } L_{\text{Grad}}^c(p) \geq T_\theta, \\
0, & \text{otherwise},
\end{cases}
\label{eq:DeGrad}
\end{equation}

\noindent where $T_\theta$ is the percentile threshold corresponding to the bottom $\theta\%$. The resulting map, $L_{\text{DeGrad}}^c$, is cleaner, more focused on the target object, and serves as a robust input for subsequent Fusion-CAM operations.

\subsection{Combination of Denoised Gradient and Region-based CAM}
\label{CAMs_integration}
To leverage the complementary strengths of gradient- and region-based CAMs, we combine them into a single class activation map (see Part~B in Figure~\ref{fig:workflow}). Gradient-based CAM provide precise, class-discriminative localization but often focus only on the most salient regions, leading to incomplete coverage. In contrast, region-based CAM offer broader spatial support with weaker boundary discrimination, ensuring more complete object coverage. By combining these two sources, the resulting map captures both high precision and comprehensive coverage of the target object.

Let $L^c_{\text{DeGrad}}$ denote the denoised gradient-based CAM for class $c$, and $L^c_{\text{Region}}$ the corresponding region-based CAM. To integrate these complementary activation maps, we first compute contribution weights, denoted as $\beta_{\text{DeGrad}}$ and $\beta_{\text{Region}}$, which quantify the relative importance of each map in predicting class $c$. These weights are obtained by applying each activation map as a spatial mask over the input image $I_0$, effectively preserving only the regions highlighted by the map, and measuring the corresponding effect on the network's class score relative to a neutral baseline provided by an all-black image $I_b$. Formally, the weights are computed as:
\vspace{-5pt}
\begin{equation}
\begin{split}
\beta_{\text{DeGrad}} = f_c(L^c_{\text{DeGrad}} \circ I_0) - f_c(I_b),\\
\beta_{\text{Region}} = f_c(L^c_{\text{Region}} \circ I_0) - f_c(I_b)
\end{split}
\label{eq: weights}
\end{equation}


\noindent where $\circ$ denotes element-wise multiplication and $f_c(\cdot)$ is the network's output score for class $c$.

Finally, the combined activation map $L^c_{\text{GradRegion}}$ is computed as a weighted linear combination of the two maps:
\vspace{-5pt}
\begin{equation}
L^c_{\text{GradRegion}} = \beta_{\text{DeGrad}} \cdot L^c_{\text{DeGrad}} + \beta_{\text{Region}} \cdot L^c_{\text{Region}}
\label{eq:Union}
\end{equation}

 This integration strategy effectively integrates the focused, class-discriminative activations of gradient-based CAM with the broader spatial support of region-based CAM, producing a more accurate and spatially complete class activation map.

\subsection{Similarity Based Fusion}
While the weighted combination of activation maps described in Section~\ref{CAMs_integration} leverages their respective contribution scores, it does not inherently guarantee enhanced localization. A key limitation arises when the denoised gradient-based map $ L^c_{\text{DeGrad}}$ retains residual background activations due to imperfect thresholding. In such cases, a disproportionately high contribution score $\beta_{\text{DeGrad}}$ can dominate the union process, causing the final map to overemphasize irrelevant or noisy regions. This imbalance becomes particularly problematic when $L^c_{\text{DeGrad}}$ covers large but semantically weak areas, whereas $ L^c_{\text{Region}} $ provides more precise yet underweighted responses. The resulting composite map may therefore exhibit diluted attention and reduced class-discriminative sharpness.
To overcome this, we introduce a \textit{similarity-based fusion} mechanism that adaptively reinforces consistent activations and suppresses conflicting ones (see Part~C in Figure~\ref{fig:workflow}). We compute an updated contribution score $\beta_{\text{GradRegion}} $ (as in Eq.~\ref{eq: weights}) for the intermediate map $L^c_{\text{GradRegion}} $, and re-weight both maps:

\vspace{-0.7cm}
\begin{equation}
\begin{split}
L_{\text{W-GradRegion}} = \beta_{\text{GradRegion}} \cdot L^c_{\text{GradRegion}},\\
L_{\text{W-Region}} = \beta_{\text{Region}} \cdot L^c_{\text{Region}}
\end{split}
\label{eq:new_weights}
\end{equation}

\noindent where $ L^c_{\text{W-GradRegion}} $ and $ L^c_{\text{W-Region}}$ denote the weighted forms of the combined and region-based maps, respectively.

Next, we estimate the spatial agreement between these two sources through a pixel-wise similarity measure $S$:
\vspace{-3pt}
\begin{equation}
\text{S}(p) = 1 - \left| L_{\text{1}}(p) - L_{\text{2}}(p) \right|
\label{eq:similarity}
\vspace{-0.1cm}\end{equation}

\noindent where \( p \) denotes a spatial position in the activation maps at pixel $p$  and $L_1(p) = L_{\text{W-GradRegion}}(p)$ and $L_2(p) = L_{\text{W-Region}}(p)$. High similarity values indicate mutual reinforcement of activations, while low similarity values reveal inconsistent or noisy regions.

The final \textit{Fusion-CAM} is then computed as a similarity-aware combination:

\vspace{-0.7cm}
\begin{equation}
L^c_{\text{Fusion-CAM}} = S \cdot \max(L_1, L_2) + \bar{S}  \cdot \tfrac{L_1 + L_2}{2}
\label{eq:fusion-cam}
\vspace{-0.1cm}\end{equation}
\noindent where $S(p) \in [0,1]$ denotes the similarity between the two activation maps  at pixel $p$, and $\bar{S}(p) = 1 - S(p)$ corresponds to the dissimilarity between the two maps.

This adaptive formulation strengthens mutually consistent activations while attenuating noisy or divergent ones, producing activation maps that are both spatially coherent and class-discriminative.
The similarity-aware fusion operates at the pixel level and adaptively blends the two activation maps based on local agreement. Specifically, when both maps agree at a given pixel location—i.e., when \( L_1(p) \approx L_2(p) \)—the similarity score \( \text{S}(p) \) approaches 1, and the fusion outputs the maximum of the two values, which serves to enhance the most confident activation while preserving agreement. This reinforcement helps highlight salient regions that are consistently emphasized by both maps. Conversely, when the maps disagree—i.e., \( L_1(p) \) and \( L_2(p) \) differ significantly—the similarity score decreases, and the fusion falls back to an average, providing a soft compromise between the conflicting activations. This ensures that neither map dominates in uncertain regions, reducing the risk of overemphasizing noisy signals.

\begin{algorithm}[h]
\caption{Fusion-CAM}
\label{alg:fusioncam_algo}
\begin{algorithmic}[1]
\Require Input image $I_0$, CNN model $f$, target class $c$, threshold $\theta$
\Ensure Final fused activation map $L^c_{\text{Fusion-CAM}}$

\State $A, W \gets f(I_0)$  \textit{// Extract feature maps and activations from the CNN}
\State $L^c_{\text{DeGrad}} \gets$ Eq.~\ref{eq:DeGrad}  \textit{// Remove weak Grad-based CAM activations}
\State $L^c_{\text{Region}} \gets$   \textit{// obtain the region-based CAM}
\State $\beta_{\text{DeGrad}}, \beta_{\text{Region}} \gets$ Eq.~\ref{eq: weights} \textit{// Get contribution weights for each map}
\State $L^c_{\text{GradRegion}} \gets$ Eq.~\ref{eq:Union}  \textit{// Combine denoised Gradient and Region-based CAMs according to their contribution weights}
\State $L^c_{\text{W-GradRegion}}, L^c_{\text{W-Region}} \gets$ Eq.~\ref{eq:new_weights} \textit{// Get weighted combined and region-based map}
\State $L^c_{\text{Fusion-CAM}} \gets$ Eq.~\ref{eq:fusion-cam}  \textit{// Adaptive pixel-wise fusion of $L^c_{\text{W-GradRegion}}$ and $L^c_{\text{W-Region}}$}
\State \Return $L^c_{\text{Fusion-CAM}}$
\end{algorithmic}
\end{algorithm}

\section{Experimental Study}
\label{experiments}

The source code used in the evaluation is available on Github\,: \url{https://anonymous.4open.science/r/Fusion-CAM-3F3B}.

\subsection{Datasets}
We conducted experiments on (1) general-purpose datasets to validate model-agnostic properties and localization performance in natural scenes, and (2) specialized datasets to test performance on fine-grained domains. This involved standard benchmarks for image classification and object recognition. To ensure representative yet computationally efficient evaluation, we randomly sampled subsets of images from each dataset. For general image classification, we selected 2,000 images in total from the ImageNet (ILSVRC2012)~\cite{imagenet} validation set and the PASCAL VOC 2007~\cite{VOC2007} test set. For plant disease detection, 1,000 images were randomly selected from the validation splits of \textit{PlantVillage}~\cite{plantVillage}, \textit{Apple Leaf Disease}, \textit{Plant Leaves}, and \textit{PlantK}, all available through the TensorFlow Datasets benchmark\,\footnote{\url{https://www.tensorflow.org/datasets/catalog}}. 

All selected images were resized to $(224 \times 224 \times 3)$, rescaled to $[0,1]$, and normalized using the standard ImageNet statistics: $\mu = [0.485, 0.456, 0.406]$ and $\sigma = [0.229, 0.224, 0.225]$. More details on the dataset characteristics can be found in Appendix~\ref{app:datasets}.

\begin{figure*}[h!]
\centering
\setlength{\tabcolsep}{2pt} 
\renewcommand{\arraystretch}{1.2} 

\begin{tabular}{c c c c c c c c c}

& \footnotesize Input & \footnotesize Grad-CAM & \footnotesize Grad-CAM++ & \footnotesize XGrad-CAM & \footnotesize Score-CAM & \footnotesize Group-CAM & \footnotesize Union-CAM & \footnotesize \textbf{Fusion-CAM} \\

    \multirow{-4}{*}{\rotatebox{90}{\footnotesize ILSV2012}} &

    \includegraphics[width=0.11\textwidth]{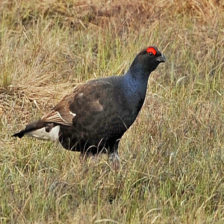} &
    \includegraphics[width=0.11\textwidth]{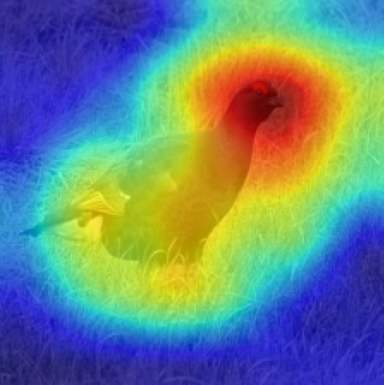} &
    \includegraphics[width=0.11\textwidth]{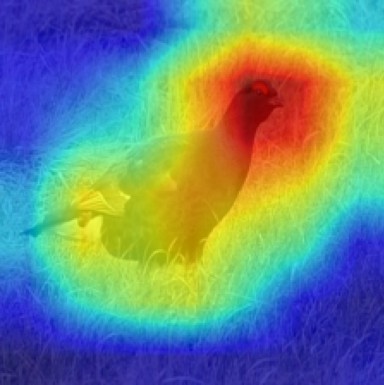} &
    \includegraphics[width=0.11\textwidth]{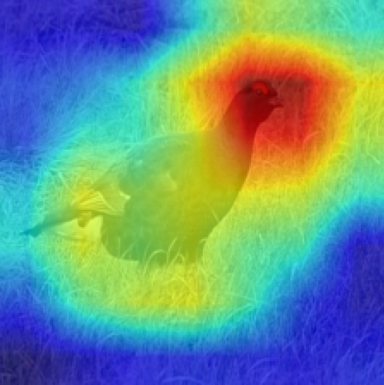} &
    \includegraphics[width=0.11\textwidth]{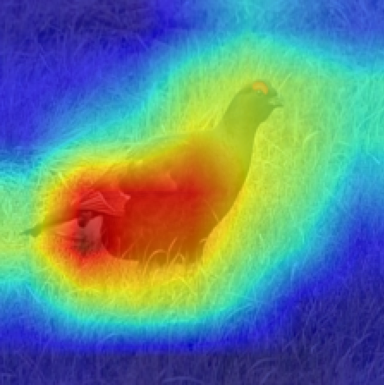} &
    \includegraphics[width=0.11\textwidth]{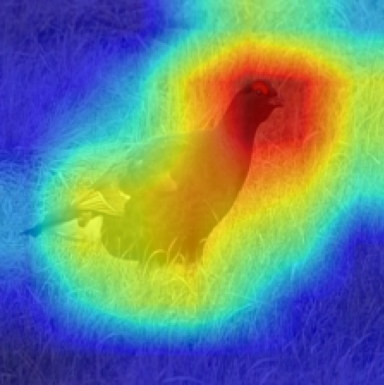} &
    \includegraphics[width=0.11\textwidth]{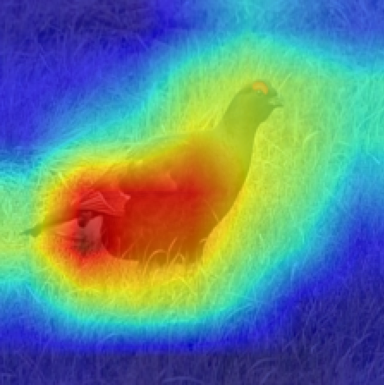} &
    \includegraphics[width=0.11\textwidth]{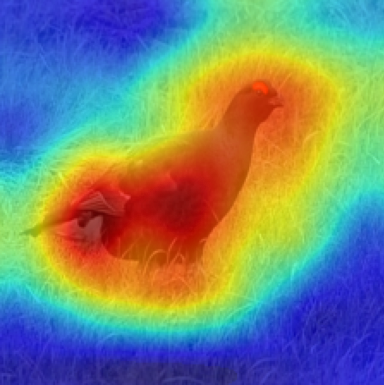} \\

    \multirow{-4}{*}{\rotatebox{90}{\footnotesize Plant Village}} &
    \includegraphics[width=0.11\textwidth]{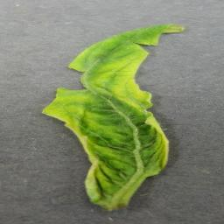} &
    \includegraphics[width=0.11\textwidth]{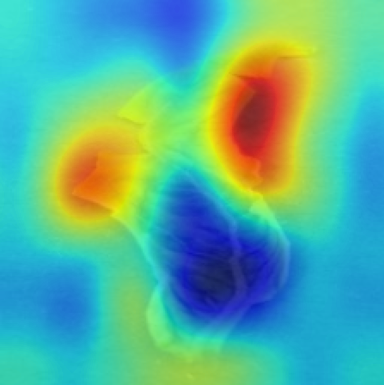} &
    \includegraphics[width=0.11\textwidth]{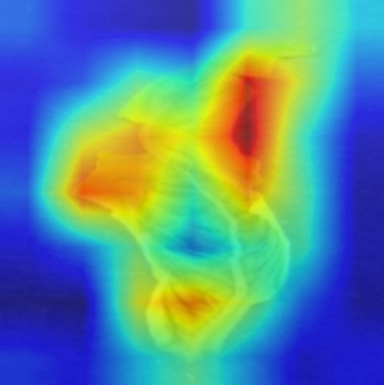} &
    \includegraphics[width=0.11\textwidth]{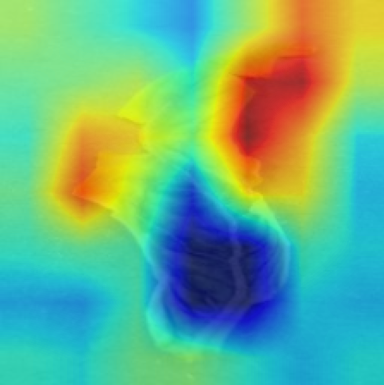} &
    \includegraphics[width=0.11\textwidth]{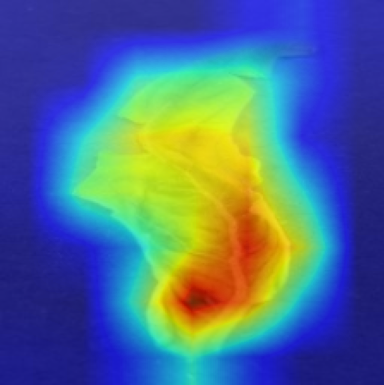} &
    \includegraphics[width=0.11\textwidth]{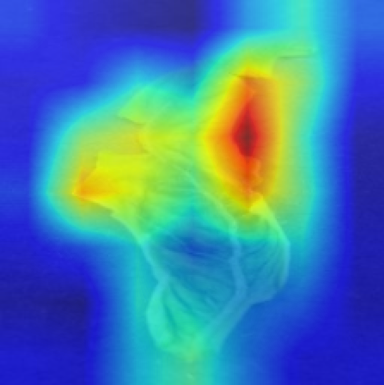} &
    \includegraphics[width=0.11\textwidth]{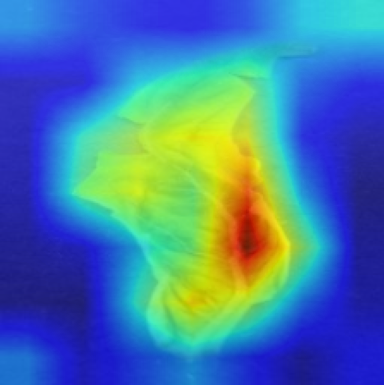} &
    \includegraphics[width=0.11\textwidth]{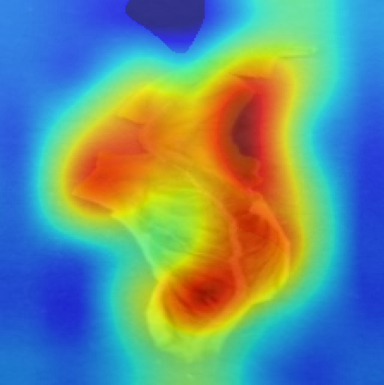} \\

    \multirow{-4}{*}{\rotatebox{90}{\footnotesize Multi-instances}} &\includegraphics[width=0.11\textwidth]{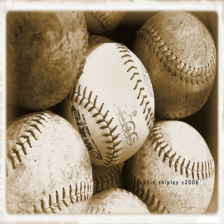} &
    \includegraphics[width=0.11\textwidth]{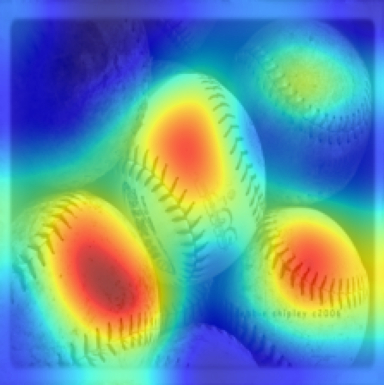} &
    \includegraphics[width=0.11\textwidth]{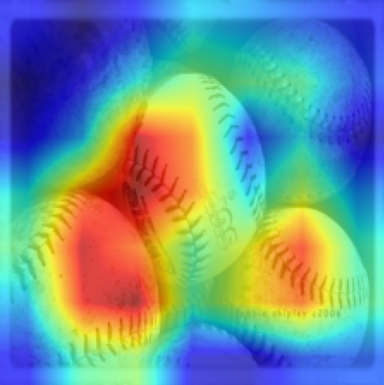} &
    \includegraphics[width=0.11\textwidth]{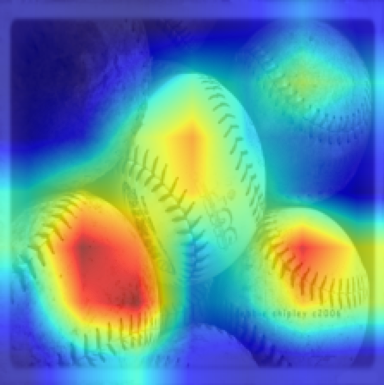} &
    \includegraphics[width=0.11\textwidth]{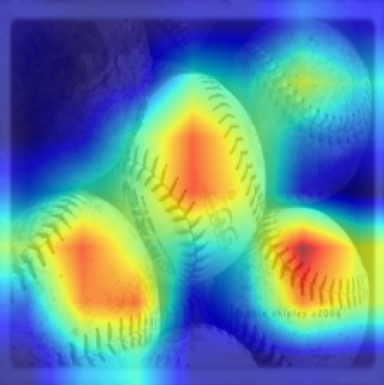} &
    \includegraphics[width=0.11\textwidth]{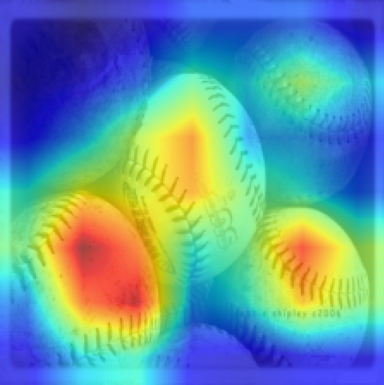} &
    \includegraphics[width=0.11\textwidth]{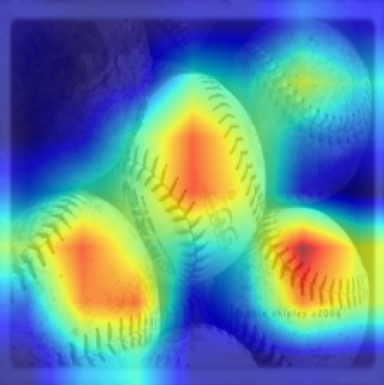} &
    \includegraphics[width=0.11\textwidth]{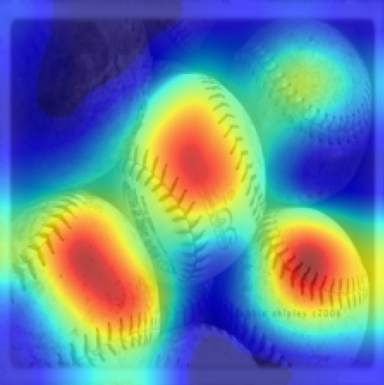} \\
  
\end{tabular}
\vspace{-0.3cm}
 \caption{Comparison of different CAM methods on examples (One instance for the two first rows, several instances for the  last).}
\label{fig:class_discriminative_Visualization}
\end{figure*}

\subsection{Base Models and Activation Maps}

To ensure architecture-agnostic evaluation, Fusion-CAM was tested on three representative convolutional backbones with different depths and parameter counts: VGG16~\cite{VGG16}, ResNet50~\cite{ResNet}, and MobileNet~\cite{MobileNet}. We report results obtained with the best-performing model, ResNet50 (50 layers, $\sim$25M parameters) (Results from other architectures are  in the Appendix~\ref{app:backbones}.
For the gradient-based component, we used Grad-CAM~\cite{Grad-CAM}, and for the region-based component, we used Score-CAM~\cite{Score-CAM}; these methods are widely adopted for CAM and provide strong baselines \cite{Union-CAM, FD-CAM} (See  Appendix~\ref{app:fusioncam_variants} for other CAM approaches). 

\subsection{Baselines and Evaluation Metrics}
We compared Fusion-CAM with several widely used class activation mapping (CAM) methods, representing different methodological paradigms and grouped by families. For Gradient-based: Grad-CAM~\cite{Grad-CAM}, Grad-CAM++~\cite{Grad-CAM++}, XGrad-CAM~\cite{XGrad-CAM}; for Region-based: Score-CAM~\cite{Score-CAM}, Group-CAM~\cite{Group-CAM} and for Ensemble Method: Union-CAM~\cite{Union-CAM}.  
All baselines were applied at inference time without modifying the model architecture.



We evaluated all methods  using both qualitative and quantitative criteria.  
For \textit{qualitative evaluation}, we visually inspected activation maps to verify whether highlighted regions align with class-specific features (Class-Discriminative) and to assess the separation of contributions from multiple classes within a single image, ensuring clear and non-overlapping explanations (Multi-Class Discriminative).
For \textit{quantitative evaluation}, we measured faithfulness using Average Drop/Increase in Confidence~\cite{Grad-CAM++}, which quantifies changes in model confidence when using only highlighted regions, and Deletion/Insertion~\cite{RISE}, which evaluates spatial relevance by progressively removing or inserting pixels based on activation maps.

\section{Results and Discussion}\vspace{-0.1cm}
\label{results}

\subsection{Fusion-Cam is effective}
We present here both qualitative and quantitative results. 

\textbf{Class Discriminative Visualization.}

 Fusion-CAM  consistently provides more precise and well-localized attention around the object of interest (See Figure~\ref{fig:class_discriminative_Visualization}). For instance  Fusion-CAM achieves more complete coverage of the entire object—the black grouse—whereas other methods focus only on specific parts, such as the top or bottom (See first row). This shows Fusion-Cam ability to capture the full spatial extent of the target.  
Moreover, Fusion-CAM remains effective even in domain-specific contexts, such as plant disease detection, where capturing fine-grained details is essential (See second row): subtle structural details like small sporadic lesions and slight discolorations are preserved.
On multiple instances (See last row), gradient-based methods such as Grad-CAM often struggle to localize them, leading to limited spatial coverage. Region-based methods like Score-CAM improve coverage but still fail to capture all relevant regions. Union-CAM enhances coverage by aggregating activation maps; however, Fusion-CAM produces more extensive activations with higher spatial coverage of the target class.
In conclusion, Fusion-CAM delivers more complete, accurate, and interpretable visual explanations. Additional visualizations illustrating diverse scenarios are provided in Appendix ~\ref{app:additionnal_visualizations}.

\textbf{Average Drop and Increase in Confidence.} 
To quantitatively evaluate the faithfulness of the visual explanations generated by Fusion-CAM for the image classification task, we adopt the evaluation metrics: \textit{Average Drop (AD)} and \textit{Increase in Confidence}, also referred to as \textit{Average Increase (AI)}, as proposed in~\cite{Grad-CAM++}. These metrics are grounded in the principle that a reliable saliency map should highlight regions responsible for the model’s prediction. Consequently for each test image, the classifier produces a baseline score \( y_i^c \) for the ground-truth class \( c \), and a masked input, obtained by element-wise multiplying the image with the Fusion-CAM map, yields a new score \( o_i^c \). The \textbf{Average Drop} (AD) is then computed as \( \text{AD} = \frac{1}{N} \sum_{i=1}^{N} \frac{\max(0, y^c_i - o^c_i)}{y^c_i} \times 100 \), penalizing saliency maps that obscure discriminative features -the lower the better. Similarly, the \textbf{Increase in Confidence} (IC) is computed as \( \text{AI} = \frac{1}{N} \sum_{i=1}^{N} \mathrm{Sign}(o^c_i > y^c_i) \times 100 \), where \( \mathrm{Sign}(\cdot) \) is the indicator function returning 1 if the condition holds and 0 otherwise, rewarding maps that highlight  salient regions most responsible for the model’s prediction -the higher, the better.
Fusion-CAM consistently achieves the lowest AD and the highest AI across all evaluated datasets (See Table~\ref{tab:AD_AI_metrics}). Across different datasets, Fusion-CAM consistently demonstrates superior localization performance. On ILSV2012, it achieves the lowest AD (13.25\%) and highest AI (42.25\%), indicating accurate focus on relevant object regions. The improvement is even larger in fine-grained plant disease detection, where AD decreases to 6.17\% and AI rises to 12.80\%. These results confirm that Fusion-CAM provides a more faithful representation of the CNN’s decision-making process compared to other methods.

\begin{table*}[h!]
\centering
\scriptsize  
\renewcommand{\arraystretch}{1.1} 
\setlength{\tabcolsep}{0.6em} 
\begin{tabular}{l|cc|cc|cc|cc|cc|cc}
\textbf{Technique} 
& \multicolumn{2}{c|}{\textbf{ILSV2012}} 
& \multicolumn{2}{c|}{\textbf{VOC2007}} 
& \multicolumn{2}{c|}{\textbf{Plant Village}} 
& \multicolumn{2}{c|}{\textbf{Plant Leaves}} 
& \multicolumn{2}{c|}{\textbf{Plant K}} 
& \multicolumn{2}{c|}{\textbf{Apple Disease}} \\
\cline{2-13}
& AD (\%) & AI (\%) & AD (\%) & AI (\%) & AD (\%) & AI (\%) & AD (\%) & AI (\%) & AD (\%) & AI (\%) & AD (\%) & AI (\%) \\
\hline
Grad-CAM     & 26.32 & 30.15 & 10.97 & 17.05 & 26.42 & 9.30  & 15.70 & 3.20  & 15.74 & 5.50  & 5.41  & 3.40  \\
Grad-CAM++   & 23.60 & 27.25 & 5.72  & 13.75 & 9.17  & 7.20  & 18.63 & 2.20  & 14.95 & 4.30  & 3.25  & 6.50  \\
XGrad-CAM    & 22.53 & 29.15 & 4.47  & 20.15 & 9.33  & 8.80  & 17.86 & 2.40  & 16.46 & 5.40  & 5.58  & 5.30  \\
Score-CAM    & 20.13 & 32.85 & 2.97  & 25.20 & 7.70  & 10.40 & 16.80 & 2.90  & 12.39 & 5.30  & 2.58  & 4.00  \\
Group-CAM    & 19.29 & 32.25 & 3.40  & 20.20 & 10.31 & 9.50  & 15.38 & 2.70  & 9.76  & 5.60  & 3.74 & 4.80  \\
Union-CAM    & 16.34 & 38.00 & 1.80  & 28.35 & 6.70  & 11.40 & 15.23 & 3.80  & 10.77 & 6.70  & 1.60  & 7.60  \\
\textbf{Fusion-CAM} & \textbf{13.25} & \textbf{42.25} & \textbf{1.56} & \textbf{29.70} & \textbf{6.17} & \textbf{12.80} & \textbf{12.43} & \textbf{4.00} & \textbf{7.75} & \textbf{7.30} & \textbf{1.04} & \textbf{8.70} \\
\end{tabular}%
\vspace{-0.3cm}
\caption{Comparative evaluation in terms of Average Drop (AD, lower is better) and Average Increase (AI, higher is better). The proposed \textbf{Fusion-CAM} consistently achieves the best performance on both metrics. Best results are in \textbf{bold}.}
\label{tab:AD_AI_metrics}

\end{table*}

\begin{table*}[h!]
\centering
\renewcommand{\arraystretch}{1.1}
\resizebox{\textwidth}{!}{%
\begin{tabular}{l|ccc|ccc|ccc|ccc|ccc|ccc} 
\textbf{ XAI Method} 
& \multicolumn{3}{c|}{\textbf{ILSV2012}} 
& \multicolumn{3}{c|}{\textbf{VOC2007}} 
& \multicolumn{3}{c|}{\textbf{Plant Village}} 
& \multicolumn{3}{c|}{\textbf{Plant Leaves}} 
& \multicolumn{3}{c|}{\textbf{Plant K}} 
& \multicolumn{3}{c|}{\textbf{Apple Disease}} \\
\cline{2-19}
& Ins. & Del. & Sc. 
& Ins. & Del. & Sc. 
& Ins. & Del. & Sc. 
& Ins. & Del. & Sc. 
& Ins. & Del. & Sc. 
& Ins. & Del. & Sc. \\
\hline
Grad-CAM     
& 61.19 & 18.23 & 42.96 
& 79.70 & 32.63 & 47.07 
& 96.71 & 65.95 & 30.76 
& 93.26 & 44.67 & 48.59 
& 90.70 & 19.64 & 71.06 
& 87.76 & 78.20 & 9.56 \\
Grad-CAM++   
& 60.08 & 20.94 & 39.14 
& 73.64 & 32.18 & 41.46 
& 96.92 & 58.97 & 37.95 
& 90.77 & 47.31 & 43.46 
& 89.84 & 20.53 & 69.31 
& 89.38 & 72.22 & 17.16 \\
XGrad-CAM    
& 64.48 & 18.30 & 46.18 
& 81.97 & 32.08 & 49.89 
& 93.25 & 51.76 & 41.49 
& 93.35 & 42.81 & 50.54 
& 91.98 & 18.86 & 73.12 
& 89.24 & 71.53 & 17.71 \\
Score-CAM    
& 64.90 & 21.13 & 43.77 
& \textbf{83.77} & 28.06 & 55.71 
& 94.44 & 47.17 & 47.27 
& 93.79 & 43.55 & 50.24 
& 92.59 & 17.68 & 74.91 
& 90.97 & 74.88 & 16.09 \\
Group-CAM    
& 65.48 & \textbf{17.86} & 47.62 
& 82.49 & 32.13 & 50.36 
& 96.85 & 54.51 & 42.34 
& 93.46 & 44.44 & 49.02 
& \textbf{92.71} & 19.59 & 73.12 
& 87.86 & 73.17 & 14.69 \\
Union-CAM    
& 65.69 & 21.08 & 44.61 
& 82.92 & 27.55 & 55.37 
& 94.59 & 50.38 & 44.21 
& 87.60 & \textbf{38.00} & 49.60 
& 92.29 & 17.72 & 74.57 
& 90.27 & \textbf{70.99}  & 19.28 \\
Fusion-CAM   
& \textbf{67.12} & 19.26 & \textbf{47.86} 
& 82.48 & \textbf{25.67} & \textbf{56.81} 
& \textbf{96.92} & \textbf{44.45} & \textbf{52.47} 
& 94.08 & 42.75 & \textbf{51.33}
& 92.09 & \textbf{16.93} & \textbf{75.16} 
& \textbf{91.52} & 71.16 & \textbf{20.36} \\
\end{tabular}%
}
\vspace{-0.3cm}
\caption{Comparative evaluation in terms of deletion (lower AUC is better) and insertion (higher AUC is better) for different 
XAI techniques  on multiple datasets. The over-all scores show that Fusion-CAM outperforms other related methods. Best results are  in \textbf{bold}.}
\label{tab:ins_del_metrics}

\end{table*}

\begin{figure*}[h!]
\centering
\setlength{\tabcolsep}{2pt} 
\renewcommand{\arraystretch}{1.2} 
\begin{tabular}{c c c c c c}
    & \footnotesize Input & \footnotesize Union-CAM & \footnotesize \footnotesize{Fusion-CAM} & \footnotesize Insertion Curves & \footnotesize Deletion Curves \\

    \multirow{-8}{*}{\rotatebox{90}{\footnotesize ILSV2012}} &
    \raisebox{0.25\height}{\includegraphics[width=0.14\textwidth]{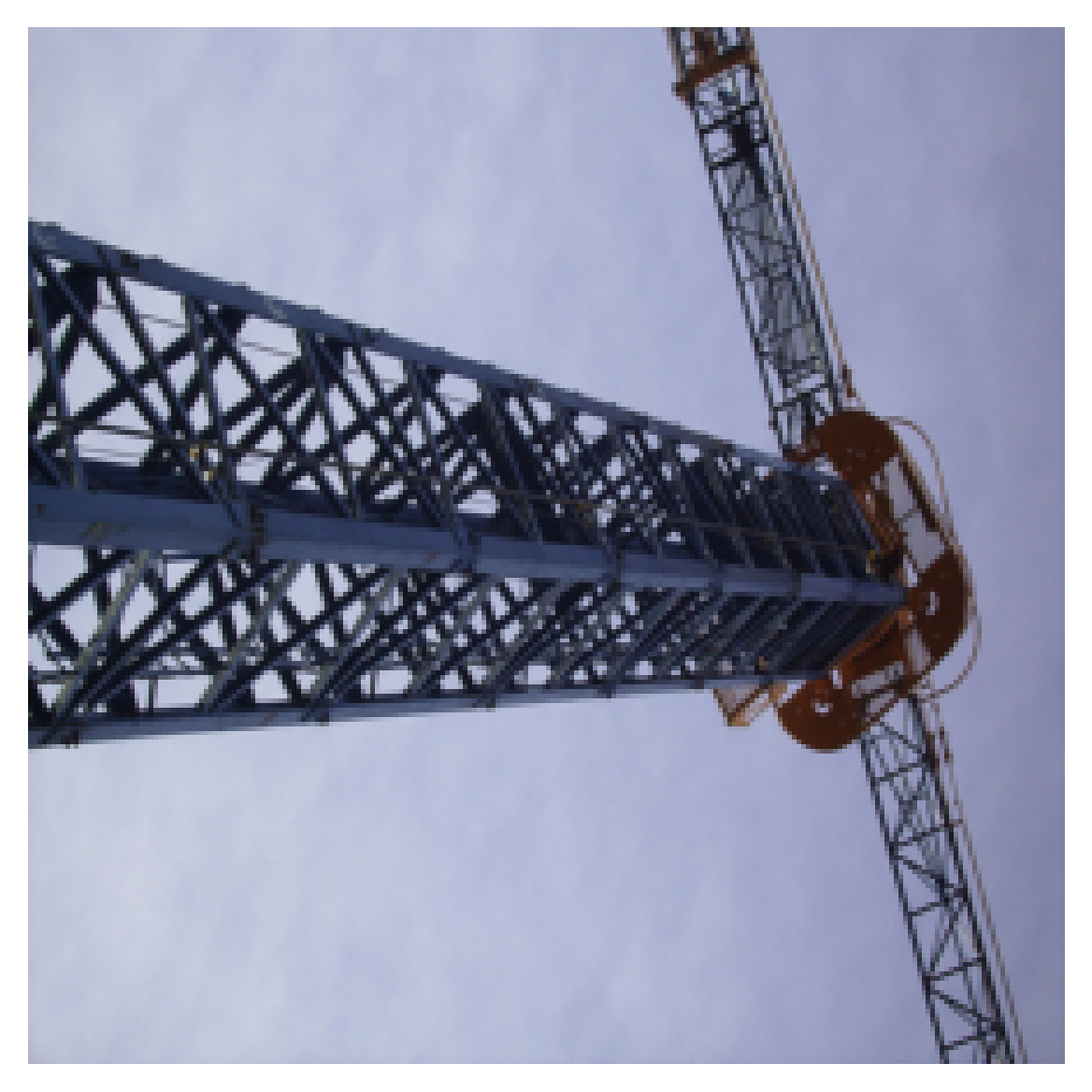}} &

    \raisebox{0.25\height}{\includegraphics[width=0.14\textwidth]{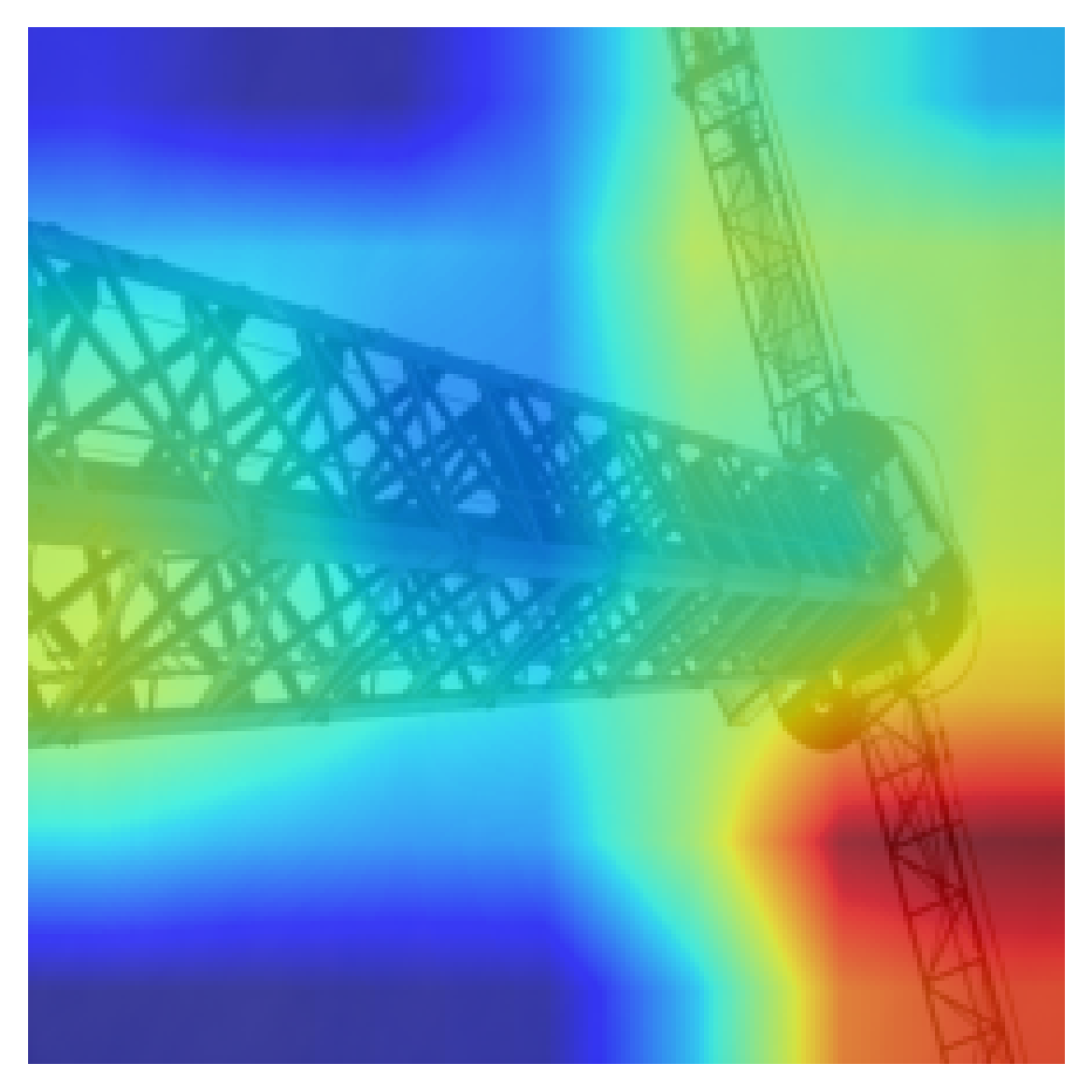}} &
    \raisebox{0.25\height}{\includegraphics[width=0.14\textwidth]{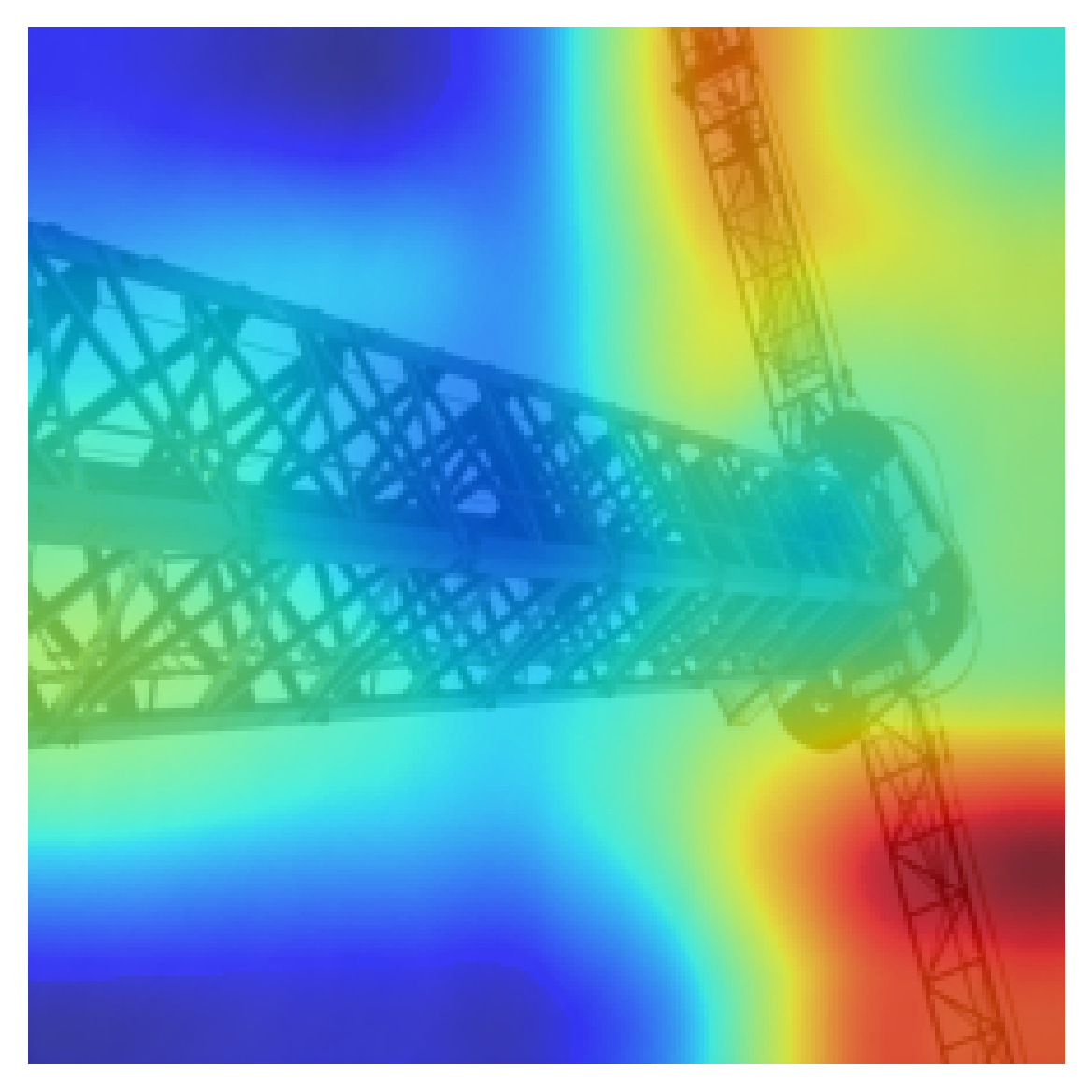}} &
    \includegraphics[width=0.24\textwidth]{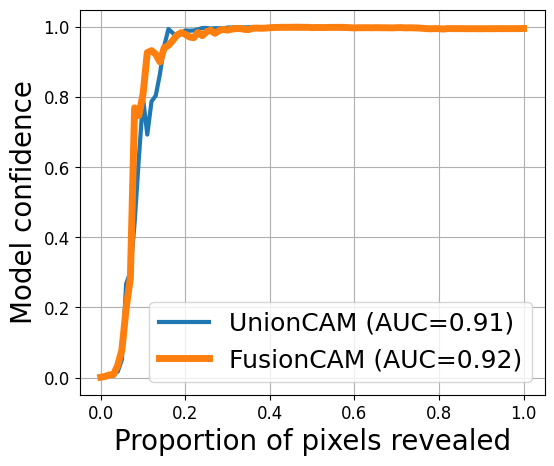} &
    \includegraphics[width=0.24\textwidth]{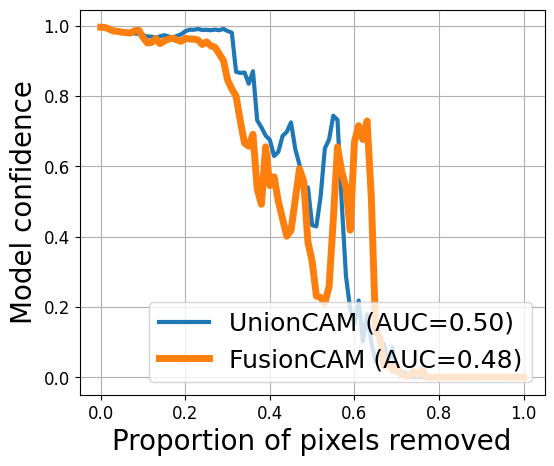} \\

    \multirow{-8}{*}{\rotatebox{90}{\footnotesize  VOC2007}} &
    \raisebox{0.25\height}{\includegraphics[width=0.14\textwidth]{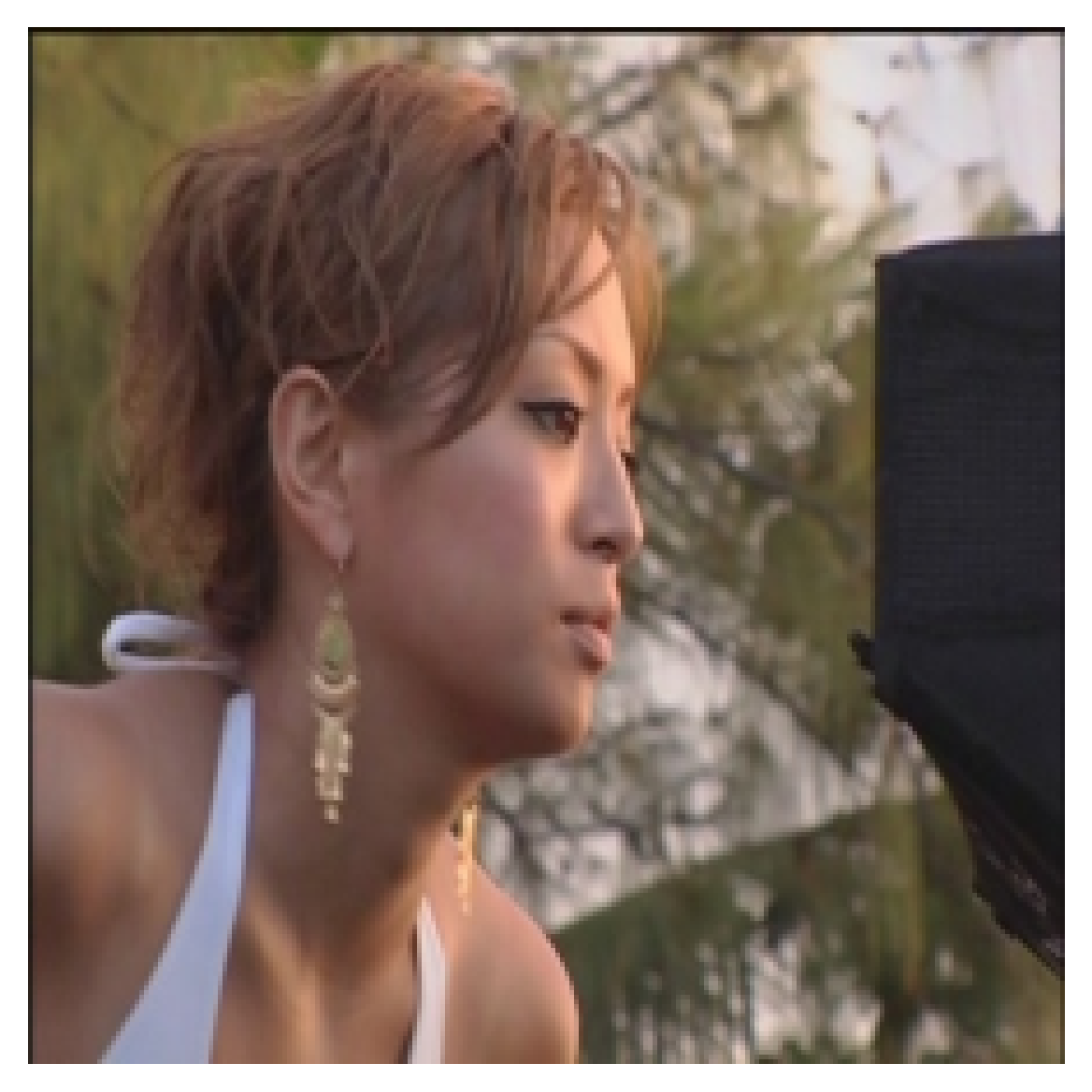}} &
    \raisebox{0.25\height}{\includegraphics[width=0.14\textwidth]{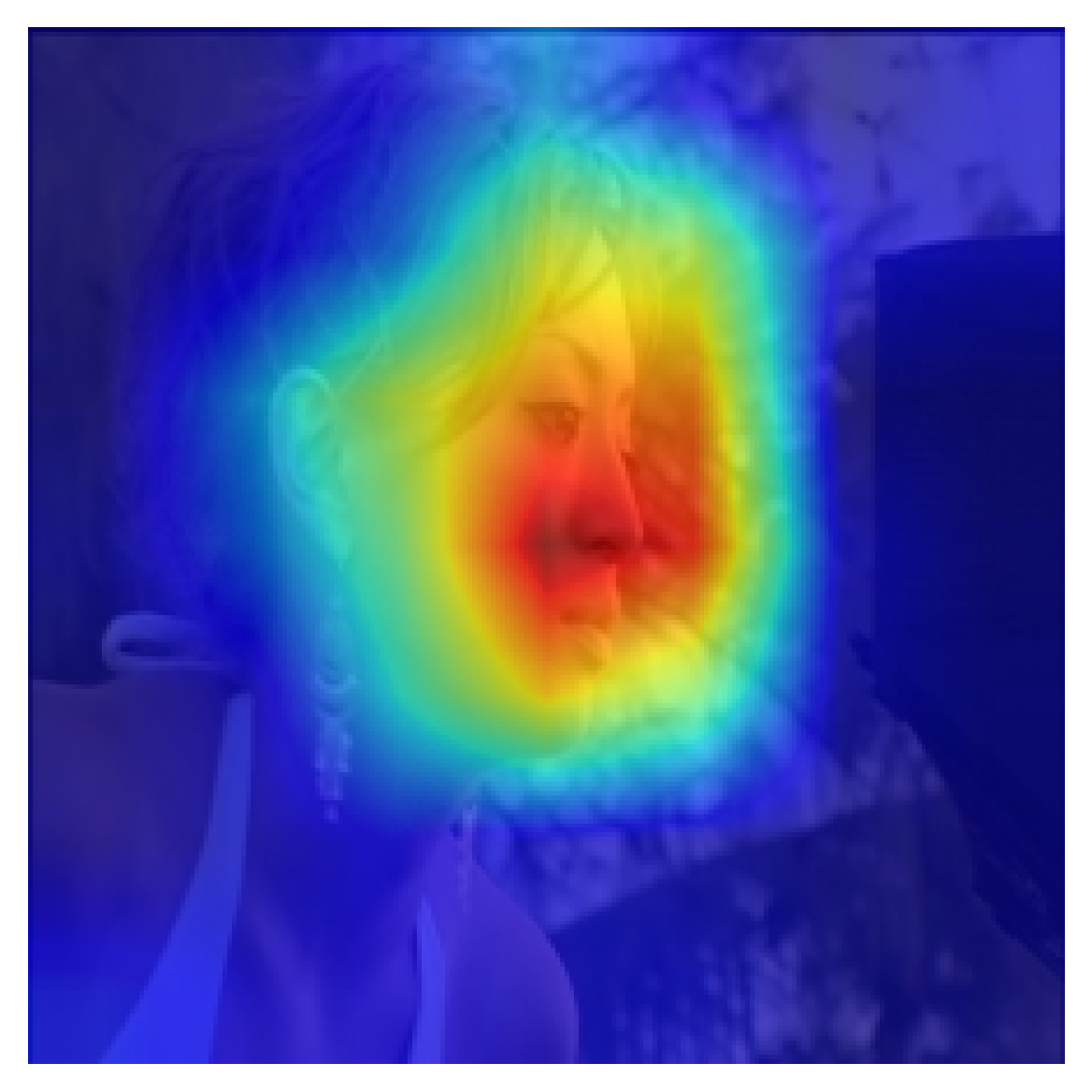}} &
    \raisebox{0.25\height}{\includegraphics[width=0.14\textwidth]{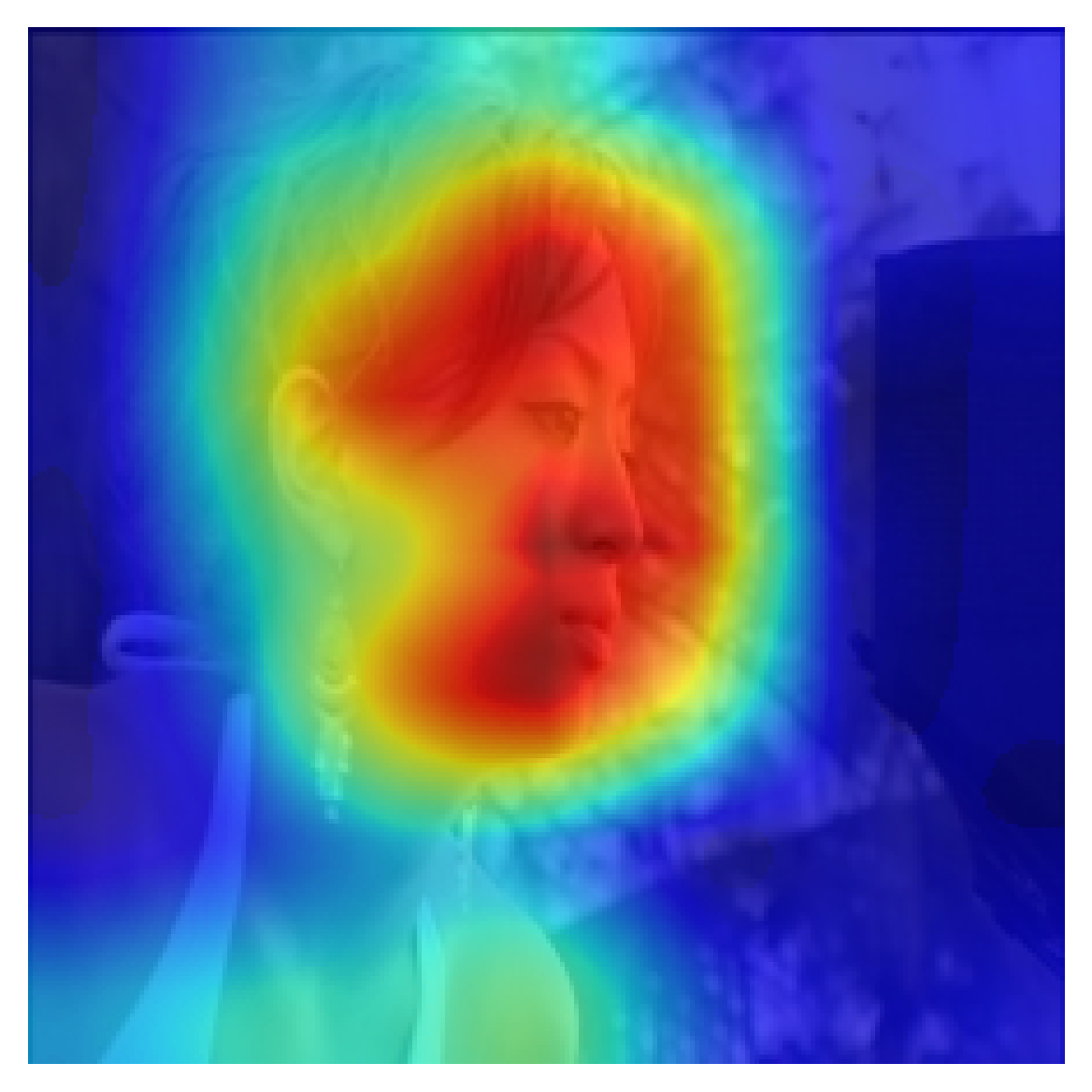}} &
    \includegraphics[width=0.24\textwidth]{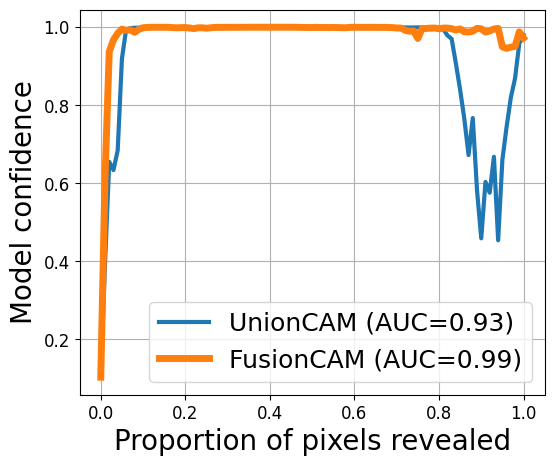} &
    \includegraphics[width=0.24\textwidth]{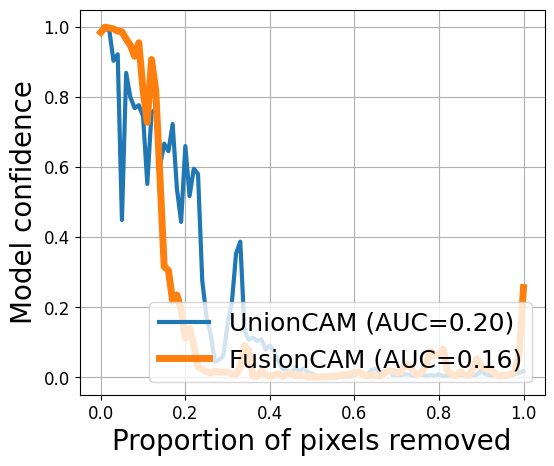} \\

\end{tabular}
\vspace{-0.3cm}
\caption{Visualization of insertion and deletion curves across multiple XAI techniques.}

\label{fig:Ins_Del_vis}
\end{figure*}

\textbf{Deletion and Insertion.}
To further evaluate the quality of the saliency maps generated by our method, we conduct experiments using the deletion and insertion metrics~\cite{RISE}. These metrics measure how model confidence changes when important pixels are removed or revealed. A lower AUC in deletion and a higher AUC in insertion indicate more faithful explanations. In our setup, pixels are modified in 1\% steps, and the Softmax AUC is used as the quantitative indicator. In addition, we report an overall score to jointly evaluate the deletion and insertion results, defined as \textit{AUC(insertion) -- AUC(deletion)}.

As reported in Table~\ref{tab:ins_del_metrics}, FusionCAM achieves the highest overall scores. The visual results in Figure ~\ref{fig:Ins_Del_vis} confirm this finding where FusionCAM produces sharper and more precise localization maps, while its insertion and deletion curves show a rapid increase and a fast drop respectively, indicating a stronger alignment between the highlighted regions and the model’s decision process. Further visualization results are presented in Appendix~\ref{app:additionnal_inser_deletion}.

\subsection{Sensitivity - Threshold Parameter $\theta$}

\begin{table*}[h!]
\footnotesize
\centering
\begin{tabular}{c|cc|cc|cc|cc|cc|cc}
$\theta$ (\%) 
& \multicolumn{2}{c|}{ILSV2012} 
& \multicolumn{2}{c|}{VOC2007} 
& \multicolumn{2}{c|}{Plant Village} 
& \multicolumn{2}{c|}{Plant Leaves} 
& \multicolumn{2}{c|}{Plant K} 
& \multicolumn{2}{c|}{Apple Disease} \\
\cline{2-13}
 & AD & AI & AD & AI & AD & AI & AD & AI & AD & AI & AD & AI \\
\hline
0   & 13.73 & 41.00 & 1.72 & 28.85 & 6.24 & 12.60 & 13.26 & 4.00 & 7.96 & 7.50 & 1.11 & 8.40 \\
10  & \textbf{13.25} & \textbf{42.25} & \textbf{1.56} & \textbf{29.70} & \textbf{6.17} & 12.80 & \textbf{12.43} & 4.00 & 7.75 & 7.30 & 1.04 & \textbf{8.70} \\
20  & 13.87 & 40.70 & 1.60 & 29.20 & 6.19 & 12.70 & 13.87 & \textbf{4.40} & \textbf{7.65} & 7.30 & \textbf{1.03} & 8.60 \\
30  & 14.81 & 39.55 & 1.70 & 28.95 & 6.20 & 12.70 & 15.13 & 3.80 & 7.72 & \textbf{7.40} & 0.22 & 6.90 \\
40  & 15.27 & 38.50 & 1.69 & 29.65 & 6.25 & \textbf{13.10} & 15.24 & 3.20 & 7.67 & 7.10 & 1.14 & 7.60 \\
50  & 15.72 & 37.45 & 1.95 & 28.65 & 6.22 & 12.60 & 16.02 & 3.00 & 7.98 & 6.90 & 1.20 & 7.20 \\
\end{tabular}
\vspace{-0.3cm}
\caption{AD and AI  across datasets for varying threshold values $\theta$. Bold values indicate the best performance for each dataset and metric.}
\label{tab:AD_AI_threshold}
\end{table*}

We evaluate the influence of  the denoising threshold parameter $\theta$\,(Sec.\ref{threshold}), varying it from 0\% to 90\%. (see Appendix~\ref{app:Threshold}).
Table~\ref{tab:AD_AI_threshold} reports Average Drop and Average Increase. It shows that a small non-zero threshold (10–20\%) provides the best trade-off between AD and AI, improving focus without removing key features. Zero thresholds increase AD and slightly lower AI, while large thresholds ($\geq 50\%$) degrade performance by discarding relevant activations. Overall, the results show that a moderate threshold, around $\theta \approx 10\%$, effectively suppresses irrelevant activations while retaining the key regions that contribute most to the model’s decision, resulting in more focused and informative activation maps.

\subsection{Efficiency}

Table~\ref{tab:temps_generation_xai} presents the computational cost of the studied XAI methods. Gradient-based methods (Grad-CAM, Grad-CAM++, XGrad-CAM) are the fastest as they require only a single backward pass. Region-based methods (Score-CAM, Group-CAM) are slower because of multiple forward passes. Ensemble approaches (Union-CAM, Fusion-CAM) are the most time-consuming. Fusion-CAM consistently outperforms Union-CAM by offering a better trade-off between computation time and explanation quality. Detailed results for the other datasets are provided in Appendix in Appendix ~\ref{app:computational_cost}.
\begin{table}[H]
\centering
\footnotesize
\begin{tabular}{l|c|c|c|c}
\textbf{XAI method} & \textbf{ILSVRC12} & \textbf{VOC07} & \textbf{Plant Village} & \textbf{AVG} \\ \hline
Grad-CAM     & 0.028 & 0.027 & 0.016  & 0.023 \\
Grad-CAM++   & 0.046 & 0.027 & 0.020  & 0.027 \\
\textbf{XGrad-CAM} & \textbf{0.027} & \textbf{0.015} & \textbf{0.015} & \textbf{0.018} \\
Score-CAM    & 4.036 & 4.085 & 1.936 & 3.378 \\
Group-CAM    & 0.498 & 0.554 & 0.368 & 0.479 \\
Union-CAM    & 5.848 & 6.140 & 2.091 & 4.562 \\
Fusion-CAM   & \underline{4.219} & \underline{4.199} & \underline{1.962} & \underline{3.478} \\
\end{tabular}
\vspace{-0.3cm}
\caption{Average generation time (s). The fastest individual methods are in \textbf{bold}, and the fastest ensemble ones are \underline{underlined}.
 }
\label{tab:temps_generation_xai}
\end{table}

\subsection{Ablation study}

We conducted a step-by-step ablation study on the ImageNet (ILSVRC2012) validation set. We designed five experimental settings in which components are progressively introduced: (1) a baseline that directly sums the Grad-CAM (without denoising) and Score-CAM without any weighting, (2) the addition of a denoising step applied to the Grad-CAM using the threshold \( \theta \), (3) the introduction of weighting coefficients \( \beta_{\text{Denoising}} \) and \( \beta_{\text{Region}} \) to control the relative importance of the Grad-CAM and Score-CAM during their union, and (4) the complete Fusion-CAM approach.

Results in Table \ref{tab:ablation_results} 
show a clear progressive improvement as Fusion-CAM components are added. Denoising modestly reduces Avg Drop and increases Avg Increase, weighted union brings larger gains, and the full Fusion-CAM achieves the best performance, highlighting the cumulative benefit of all components.

\vspace{-5pt}
\begin{table}[H]
\centering
\footnotesize

\begin{tabular}{c|l|c|c}
\textbf{Setting} & \textbf{Description} & \makecell{\textbf{Average} \\ \textbf{Drop (\%)}} & \makecell{\textbf{Average} \\ \textbf{Increase (\%)}} \\
\hline
1 & Baseline & 17.96 & 36.75 \\
2 & Baseline + Denoising \( \theta \) (10\%) & 17.10 & 36.95 \\
3 & Baseline + Weighted Union & 15.26 & 38.55 \\
4 & Full pipeline & \textbf{13.25} & \textbf{42.25} \\
\end{tabular}
\vspace{-0.3cm}
\caption{Ablation study  on ImageNet (ILSVRC2012) validation set, showing the impact of progressively adding denoising, weighted union, and similarity-based fusion on  AD and AI.}
\label{tab:ablation_results}
\end{table}

\section{Conclusion}
\label{sec:conclusion}

We introduced Fusion-CAM, a novel framework for generating visual explanations from deep convolutional neural networks. By leveraging the complementary strengths of gradient and region-based CAM methods, we proposed a fusion strategy that integrates denoised Grad-CAM with Score-CAM through confidence-weighted aggregation and a pixel-wise similarity-based fusion mechanism. Our extensive experimental evaluation across multiple benchmark datasets—including ILSVRC2012 (ImageNet), PASCAL VOC, and several plant disease datasets—demonstrates the consistent superiority of Fusion-CAM over existing state-of-the-art methods.  

Qualitatively, Fusion-CAM produces saliency maps that are both discriminative and broader contextual regions. Quantitatively, it achieves the best performance on standard evaluation metrics, such as the lowest Average Drop, the highest Average Increase in Confidence, and optimal overall scores on insertion and deletion metrics, indicating its faithfulness in identifying the regions most critical for the model's predictions. Furthermore, ablation studies validated the contribution of each component in our pipeline in enhancing the final visual explanation map.  
Overall, Fusion-CAM provides a robust and effective tool for model interpretability. Beyond image classification, its fusion paradigm opens promising directions for emerging architectures like Vision Transformers, where understanding the decision-making process is crucial for safely and effectively deploying such models in real-world applications.


{
    \small
    \bibliographystyle{ieeenat_fullname}
    \bibliography{main}
}

\appendix

\clearpage
\setcounter{page}{1}
\maketitlesupplementary

This supplementary material provides additional experiments and analyses supporting the findings presented in the main paper. It includes more dataset descriptions, comparisons across different backbone architectures, supplementary qualitative visualizations, insertion and deletion analyses, computational cost evaluations, evaluation using other Gradient-based CAM variants within Fusion-CAM, assessing Fusion-CAM robustness and generalization across diverse datasets and model architectures.

\section{Datasets }
\label{app:datasets}
Table~\ref{tab:datasets} summarises the characteristics of the subset of datasets used in this study. In the literature, the number of samples used for evaluating Class Activation Map (CAM) techniques varies: some studies rely on 1000 random images~\cite{GradPPScoreCAM}, while the majority follow the standard practice of using 2000 random samples, particularly for large-scale datasets such as ILSVRC2012 and VOC2007~\cite{Union-CAM,FD-CAM,Increment-CAM}. As indicated in the main paper and following this standard, we selected 2000 random samples from the ILSVRC2012 validation set (ImageNet) and the VOC2007 test set; this choice ensures that our results remain directly comparable to the existing literature. For the plant disease datasets, we limited the evaluation to 1000 samples, since increasing the sample size to 2000 does not affect the relative differences between the CAM techniques.

\begin{table}[htbp]
\centering
\scriptsize
\setlength{\tabcolsep}{4pt}
\renewcommand{\arraystretch}{1.2}
\begin{tabular}{|>{\centering\arraybackslash}m{2.2cm}|>{\centering\arraybackslash}m{1.0cm}|>{\arraybackslash}m{3.8cm}|}
\hline
\textbf{Dataset} & \textbf{Images} & \textbf{Description} \\
\hline
ILSVRC 2012 (Val.) & \multirow{2}{*}{2000} & Large-scale benchmark for natural image classification. \\
\cline{1-1}\cline{3-3}
PASCAL VOC 2007 &  & Object detection and classification benchmark with multi-object scenes. \\
\hline
PlantVillage & \multirow{4}{*}{1000} & Extensive plant leaf dataset (healthy and diseased conditions). \\
\cline{1-1}\cline{3-3}
Plant Leaves &  & Leaf images categorized by species and health state. \\
\cline{1-1}\cline{3-3}
Apple Leaf Disease &  & Apple leaf disease dataset: Scab, Black Rot, Cedar Apple Rust, Healthy. \\
\cline{1-1}\cline{3-3}
PlantK &  & High-resolution leaves annotated by species and health status. \\
\hline
\end{tabular}
\caption{Overview of the data used for evaluation.}
\label{tab:datasets}
\end{table}


\vspace{-0.3cm}
\section{Evaluation Across Backbone Architectures}
\label{app:backbones}

To ensure that Fusion-CAM is architecture-agnostic, it was tested on three representative convolutional backbones with different depths and parameter counts: VGG16~\cite{VGG16} (16 layers, 138M parameters), ResNet50~\cite{ResNet} (50 layers, 25.6M parameters), and MobileNet~\cite{MobileNet} (28 layers, 4.2M parameters). All three models achieve strong and consistent results, with ResNet50 slightly outperforming the others when evaluated on the validation sets of the 6 considered datasets. For example, on the VOC 2007 dataset, ResNet50 achieved an accuracy of 96.64\%, compared to 96.13\% for VGG16 and 77.19\% for MobileNet. Similarly, on the Plant Leaves dataset, ResNet50 reached 98.50\%, VGG16 96.50\% and MobileNet 90.00\%, showing the same tendency across the rest of the datasets. While the main paper reports results using ResNet50, we evaluated FusionCAM with VGG16 and MobileNet to further assess its robustness across different architectures. We report the  \textit{Average Drop (AD)} and \textit{Average Increase (AI)} metrics for VGG16 in Tables~\ref{tab:AD_AI_vgg16} and for MobileNet in Table~\ref{tab:AD_AI_mobilnet}. Consistent with the findings reported for ResNet50 in the main paper, FusionCAM also achieves the lowest AD and highest AI on both VGG16 and MobileNet architectures, confirming its ability to produce faithful visual explanations regardless of the underlying CNN architecture.

\begin{table*}[h!]
\centering
\footnotesize
\renewcommand{\arraystretch}{1.1} 
\setlength{\tabcolsep}{0.6em} 
\resizebox{\textwidth}{!}{%
\begin{tabular}{|l|cc|cc|cc|cc|cc|cc|}
\hline
\textbf{Technique} 
& \multicolumn{2}{c|}{\textbf{ILSV2012}} 
& \multicolumn{2}{c|}{\textbf{VOC2007}} 
& \multicolumn{2}{c|}{\textbf{Plant Village}} 
& \multicolumn{2}{c|}{\textbf{Plant Leaves}} 
& \multicolumn{2}{c|}{\textbf{Plant K}} 
& \multicolumn{2}{c|}{\textbf{Apple Disease}} \\
\cline{2-13}
& AD (\%) & AI (\%) & AD (\%) & AI (\%) & AD (\%) & AI (\%) & AD (\%) & AI (\%) & AD (\%) & AI (\%) & AD (\%) & AI (\%) \\
\hline
Grad-CAM     & 16.88 & 37.20 & 30.92 & 31.00 & 76.86 & 5.00  & 1.85 & 33.00  & 45.70 & 16.67  & 1.06  & 35.50  \\
Grad-CAM++   & 21.10 & 32.25 & 26.96  & 29.00 & 72.23  & 8.00  & 0.69 & 66.00  & 40.18 & 35.00  & 1.34  & 32.00  \\
XGrad-CAM    & 18.15 & 35.00 & 26.96  & 29.00 & 71.50  & 8.00  & 0.73 & 64.00  & 50.80 & 8.67  & 1.09  & 34.50  \\
Score-CAM    & 16.00 & 39.85 & 18.93  & \textbf{43.00}  & 62.62  & 10.00 & 0.60 & 70.00  & 33.58 & 43.00  & 0.90  & 36.00  \\
Group-CAM    & 16.23 & 37.70 & 25.99  & 31.00 & 71.97 & 9.00  & 0.66 & 68.00  & 42.09  & 16.33  & 1.11 & 36.00  \\
Union-CAM    & 15.03 & 40.50 & 18.74  & \textbf{43.00} & 61.99  & 10.00 & 0.59 & 70.50  & 35.36 & 44.33  & 0.74  & 39.00  \\
\textbf{Fusion-CAM} & \textbf{11.49} & \textbf{43.10} & \textbf{17.82} & \textbf{43.00} & \textbf{58.35} & \textbf{11.00} & \textbf{0.56} & \textbf{73.50} & \textbf{29.79} & \textbf{46.67} & \textbf{0.44} & \textbf{41.50} \\
\hline
\end{tabular}}%
\caption{Average Drop (AD) and Average Increase (AI) using the VGG16 classifier. Fusion-CAM  achieves the best performance on both metrics. Best results are in \textbf{bold}.}

\label{tab:AD_AI_vgg16}

\end{table*}

\begin{table*}[h!]
\centering
\renewcommand{\arraystretch}{1.1} 
\setlength{\tabcolsep}{0.6em} 
\resizebox{\textwidth}{!}{%
\begin{tabular}{|l|cc|cc|cc|cc|cc|cc|}
\hline
\textbf{Technique} 
& \multicolumn{2}{c|}{\textbf{ILSV2012}} 
& \multicolumn{2}{c|}{\textbf{VOC2007}} 
& \multicolumn{2}{c|}{\textbf{Plant Village}} 
& \multicolumn{2}{c|}{\textbf{Plant Leaves}} 
& \multicolumn{2}{c|}{\textbf{Plant K}} 
& \multicolumn{2}{c|}{\textbf{Apple Disease}} \\
\cline{2-13}
& AD (\%) & AI (\%) & AD (\%) & AI (\%) & AD (\%) & AI (\%) & AD (\%) & AI (\%) & AD (\%) & AI (\%) & AD (\%) & AI (\%) \\
\hline
Grad-CAM     & 47.98 & 14.00 & 37.67 & 25.00 & 40.65 & 36.50  & 43.53 & 5.00  & 22.52 & 14.00  & 42.57  & 18.00  \\
Grad-CAM++   & 47.03 & 15.00 & 36.35  & 27.00 & 31.45  & 44.50 & 30.93 & 10.00  & 20.88 & 21.00  & 32.72  & 21.00  \\
XGrad-CAM    & 44.51 & 17.00 & 34.79  & 29.00 & 30.36  & 49.00  & 28.80 & 25.00  & 18.56 & 24.00  & 29.56  & 22.00  \\
Score-CAM    & 30.95 & 24.00 & 32.72  & 34.00 & 18.41  & 64.00 & 17.12 & 30.00  & 13.57 & 32.00  & 23.21  & 38.00  \\
Group-CAM    & 46.02 & 15.00 & 33.05  & 30.00 & 27.69 & 49.50  & 17.46 & 28.00  & 19.10  & 27.00  & 27.63 & 29.00  \\
Union-CAM    & 30.64 & 24.00 & 32.64  & 34.00 & 17.45  & 65.50 & 16.86 & 30.00  & 13.33 & 33.00  & 23.47 & 37.00  \\
\textbf{Fusion-CAM} & \textbf{29.60} & \textbf{26.00} &\textbf{30.61} & \textbf{37.00} & \textbf{14.53} & \textbf{70.00} & \textbf{13.08} & \textbf{35.00} & \textbf{12.87} & \textbf{35.00} & \textbf{22.76} & \textbf{39.00} \\  
\hline
\end{tabular}}%
\caption{Average Drop (AD) and Average Increase (AI) using the MobileNet classifier. Fusion-CAM achieves the best performance on both metrics. Best results are in \textbf{bold}.}
\label{tab:AD_AI_mobilnet}

\end{table*}

\section{Fusion-CAM Variants}
\label{app:fusioncam_variants}
We conducted other experiments to evaluate whether replacing the gradient-based component in Fusion-CAM with more advanced variants could further enhance performance. Specifically, we examined Grad-CAM++ which improves multi-instance localization through higher-order derivatives and XGrad-CAM, which offers refined sensitivity to fine-grained features through gradient-normalized weighting. While these standalone methods outperform Grad-CAM in terms of localization and detail preservation, the results in  table~\ref{tab:fusioncam_variants} reveal an interesting outcome: when these variants are integrated into the Fusion process, the version built on the original Grad-CAM consistently delivers better overall performance. This suggests that Grad-CAM's simpler gradient computation provides a more stable foundation for fusion, it interacts more effectively with the region based signals, leading to a more robust attention map compared to using its variants.


\begin{table*}[h!]
\centering
\renewcommand{\arraystretch}{1.2} 
\setlength{\tabcolsep}{1.2em}     
\Huge                           
\resizebox{\textwidth}{!}{%
\begin{tabular}{|l|cc|cc|cc|cc|cc|cc|}
\hline
\textbf{Fusion-CAM variants} 
& \multicolumn{2}{c|}{\textbf{ILSVRC2012}} 
& \multicolumn{2}{c|}{\textbf{VOC2007}} 
& \multicolumn{2}{c|}{\textbf{Plant Village}} 
& \multicolumn{2}{c|}{\textbf{Plant Leaves}} 
& \multicolumn{2}{c|}{\textbf{Plant K}} 
& \multicolumn{2}{c|}{\textbf{Apple Disease}} \\
\cline{2-13}
& AD (\%) & AI (\%) & AD (\%) & AI (\%) & AD (\%) & AI (\%) & AD (\%) & AI (\%) & AD (\%) & AI (\%) & AD (\%) & AI (\%) \\
\hline
Fusion-CAM (GradCAM, ScoreCAM)     & \textbf{13.25} & \textbf{42.25} & \textbf{1.56} & \textbf{29.70} & 6.17 & \textbf{12.80} & \textbf{12.43} & 4.00 & \textbf{7.75} & \textbf{7.30} & \textbf{1.04} & 8.70 \\
Fusion-CAM (GradCAM++, ScoreCAM)   & 14.77 & 37.55 & 2.01 & 22.95 & \textbf{5.11} & 11.80 & 12.79 & \textbf{5.40} & 9.82 & 6.60 & 1.31 & \textbf{10.70} \\
Fusion-CAM (XGradCAM, ScoreCAM) & 14.91 & 38.10 & 1.86 & 26.80 & 5.93 & 11.70 & 12.84 & 5.10 & 9.85 & 6.60 & 1.22 & 8.30 \\ 
\hline
\end{tabular}%
}
\caption{Fusion-CAM variants across multiple datasets in terms of Average Drop (AD) and Average Increase (AI). While Grad-CAM++, and XGrad-CAM improve upon their baselines, the \textit{Fusion-CAM} variant built on Grad-CAM consistently achieves the best performance. Best results are in \textbf{bold}.}
\label{tab:fusioncam_variants}
\end{table*}

\section{Additionnal Visual Evaluation}
\label{app:additionnal_visualizations}

\begin{figure*}[h!]
\centering
\setlength{\tabcolsep}{2pt} 
\renewcommand{\arraystretch}{1.2} 

\begin{tabular}{c c c c c c c c c}

& \footnotesize Input & \footnotesize Grad-CAM & \footnotesize Grad-CAM++ & \footnotesize XGrad-CAM & \footnotesize Score-CAM & \footnotesize Group-CAM & \footnotesize Union-CAM & \footnotesize \textbf{Fusion-CAM} \\

    \multirow{-4}{*}{\rotatebox{90}{\footnotesize ILSV2012}} &

    \includegraphics[width=0.11\textwidth]{fig/ILSV2012_examples/Black_grouse/ILSVRC2012_val_00000885_input.png} &
    \includegraphics[width=0.11\textwidth]{fig/ILSV2012_examples/Black_grouse/ILSVRC2012_val_00000885_gradcam_pred_black_grouse.png} &
    \includegraphics[width=0.11\textwidth]{fig/ILSV2012_examples/Black_grouse/ILSVRC2012_val_00000885_gradcampp_pred_black_grouse.png} &
    \includegraphics[width=0.11\textwidth]{fig/ILSV2012_examples/Black_grouse/ILSVRC2012_val_00000885_xgradcam_pred_black_grouse.png} &
    \includegraphics[width=0.11\textwidth]{fig/ILSV2012_examples/Black_grouse/ILSVRC2012_val_00000885_scorecam_pred_black_grouse.png} &
    \includegraphics[width=0.11\textwidth]{fig/ILSV2012_examples/Black_grouse/ILSVRC2012_val_00000885_groupcam_pred_black_grouse.png} &
    \includegraphics[width=0.11\textwidth]{fig/ILSV2012_examples/Black_grouse/ILSVRC2012_val_00000885_unioncam_pred_black_grouse.png} &
    \includegraphics[width=0.11\textwidth]{fig/ILSV2012_examples/Black_grouse/ILSVRC2012_val_00000885_fusioncam_pred_black_grouse.png} \\

    \multirow{-4}{*}{\rotatebox{90}{\footnotesize VOC2007}} &
    \includegraphics[width=0.11\textwidth]{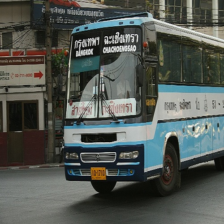} &
    \includegraphics[width=0.11\textwidth]{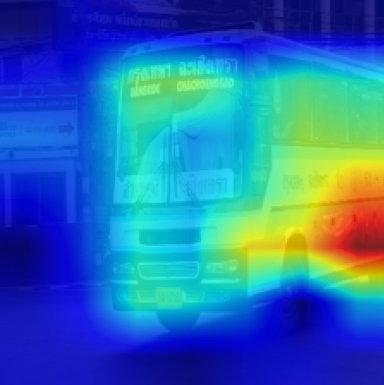} &
    \includegraphics[width=0.11\textwidth]{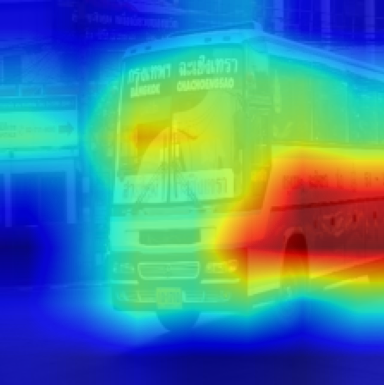} &
    \includegraphics[width=0.11\textwidth]{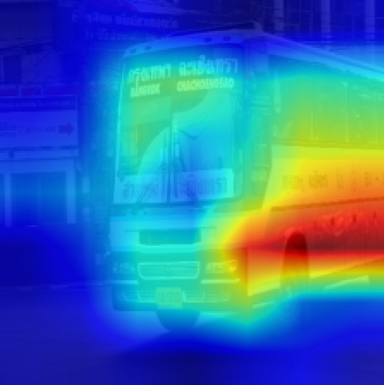} &
    \includegraphics[width=0.11\textwidth]{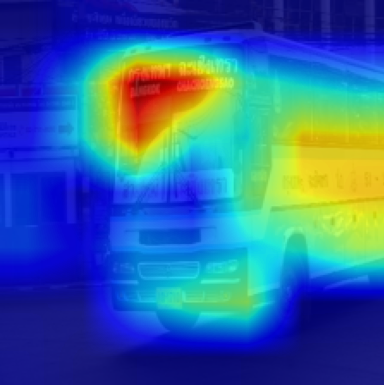} &
    \includegraphics[width=0.11\textwidth]{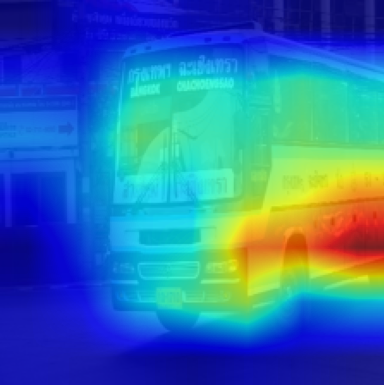} &
    \includegraphics[width=0.11\textwidth]{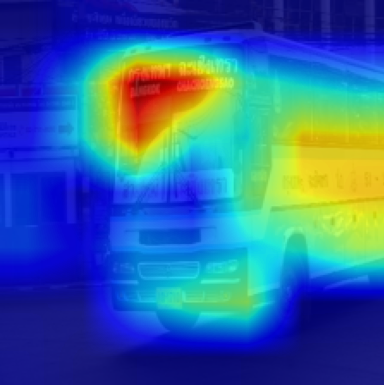} &
    \includegraphics[width=0.11\textwidth]{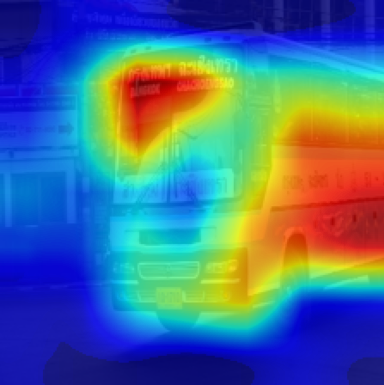} \\

    \multirow{-4}{*}{\rotatebox{90}{\footnotesize Plant Village}} &
    \includegraphics[width=0.11\textwidth]{fig/PlantVillage/6d25a71f-4e06-4a39-ac71-66fac0d23fb3___UF_input.png} &
    \includegraphics[width=0.11\textwidth]{fig/PlantVillage/6d25a71f-4e06-4a39-ac71-66fac0d23fb3___UF_gradcam_pred_Tomato__Tomato_YellowLeaf__Curl_Virus.png} &
    \includegraphics[width=0.11\textwidth]{fig/PlantVillage/6d25a71f-4e06-4a39-ac71-66fac0d23fb3___UF_gradcampp_pred_Tomato__Tomato_YellowLeaf__Curl_Virus.png} &
    \includegraphics[width=0.11\textwidth]{fig/PlantVillage/6d25a71f-4e06-4a39-ac71-66fac0d23fb3___UF_xgradcam_pred_Tomato__Tomato_YellowLeaf__Curl_Virus.png} &
    \includegraphics[width=0.11\textwidth]{fig/PlantVillage/6d25a71f-4e06-4a39-ac71-66fac0d23fb3___UF_scorecam_pred_Tomato__Tomato_YellowLeaf__Curl_Virus.png} &
    \includegraphics[width=0.11\textwidth]{fig/PlantVillage/6d25a71f-4e06-4a39-ac71-66fac0d23fb3___UF_groupcam_pred_Tomato__Tomato_YellowLeaf__Curl_Virus.png} &
    \includegraphics[width=0.11\textwidth]{fig/PlantVillage/6d25a71f-4e06-4a39-ac71-66fac0d23fb3___UF_unioncam_pred_Tomato__Tomato_YellowLeaf__Curl_Virus.png} &
    \includegraphics[width=0.11\textwidth]{fig/PlantVillage/6d25a71f-4e06-4a39-ac71-66fac0d23fb3___UF_fusioncam_pred_Tomato__Tomato_YellowLeaf__Curl_Virus.png}  \\

     \multirow{-4}{*}{\rotatebox{90}{\footnotesize Plant Leaves}} &
     \includegraphics[width=0.11\textwidth]{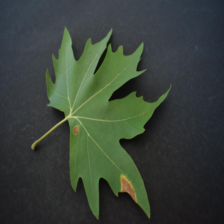} &
     \includegraphics[width=0.11\textwidth]{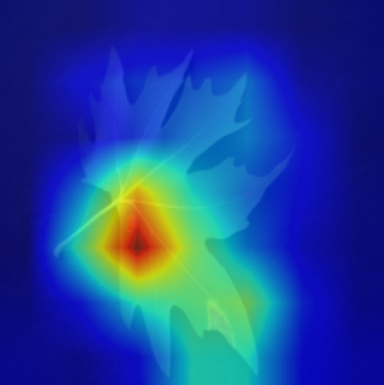} &
     \includegraphics[width=0.11\textwidth]{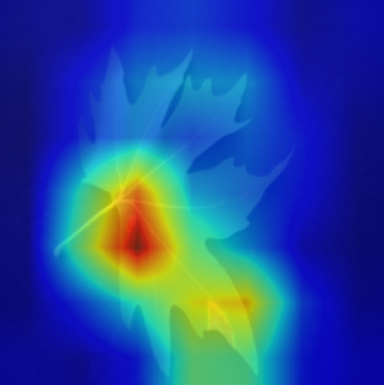} &
     \includegraphics[width=0.11\textwidth]{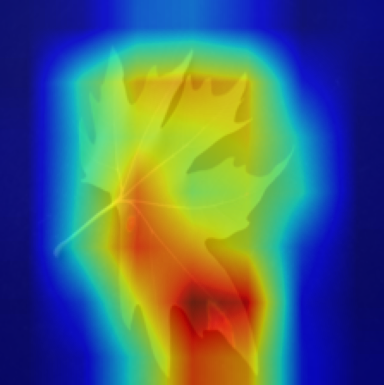} &
     \includegraphics[width=0.11\textwidth]{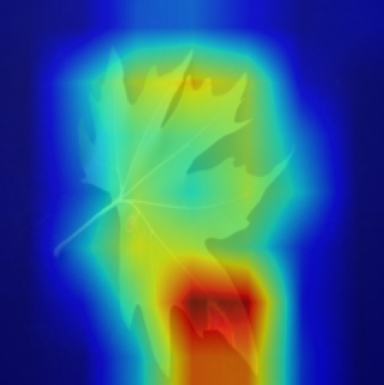} &
     \includegraphics[width=0.11\textwidth]{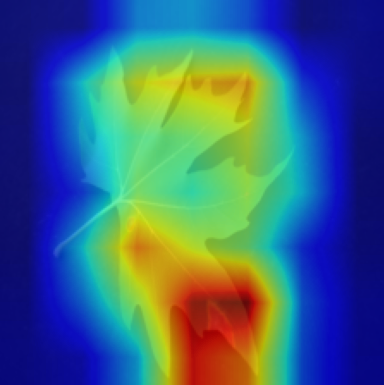} &
     \includegraphics[width=0.11\textwidth]{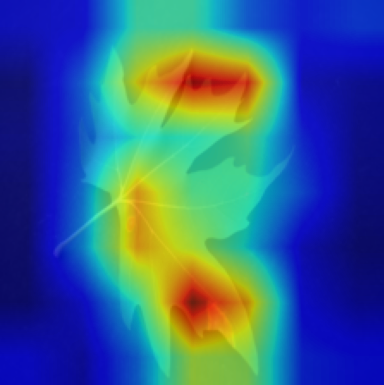} &
     \includegraphics[width=0.11\textwidth]{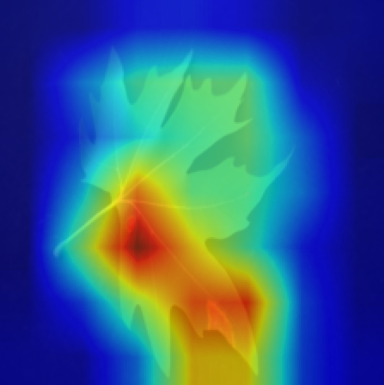} \\

     \multirow{-4}{*}{\rotatebox{90}{\footnotesize Plant K}} &
     \includegraphics[width=0.11\textwidth]{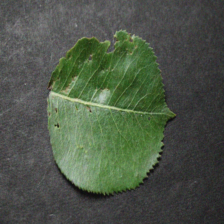} &
     \includegraphics[width=0.11\textwidth]{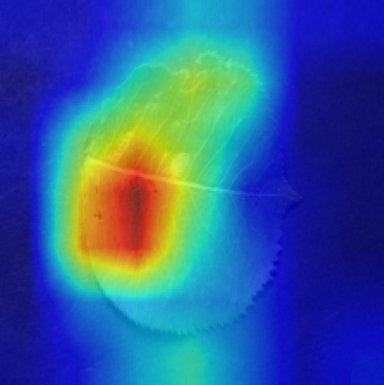} &
     \includegraphics[width=0.11\textwidth]{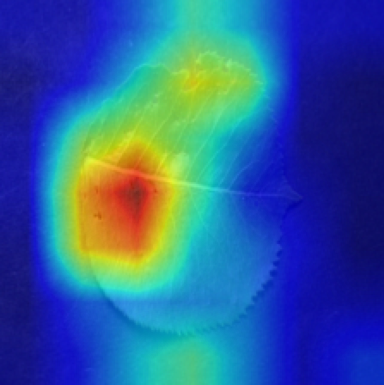} &
     \includegraphics[width=0.11\textwidth]{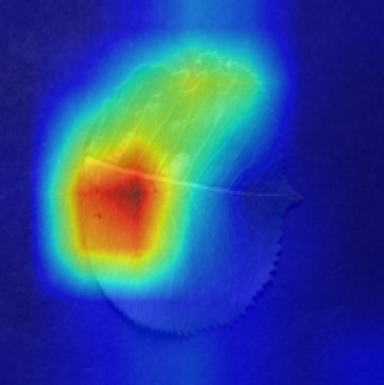} &
     \includegraphics[width=0.11\textwidth]{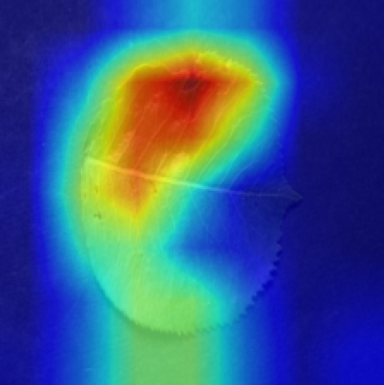} &
     \includegraphics[width=0.11\textwidth]{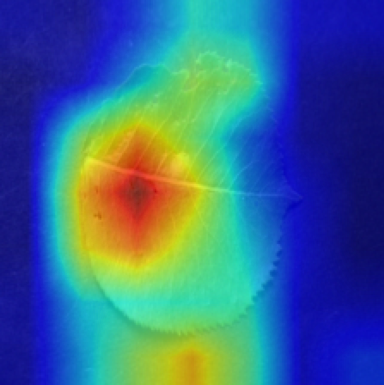} &
     \includegraphics[width=0.11\textwidth]{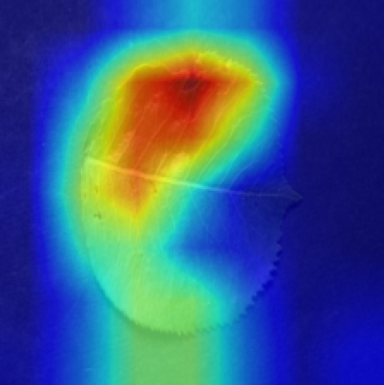} &
     \includegraphics[width=0.11\textwidth]{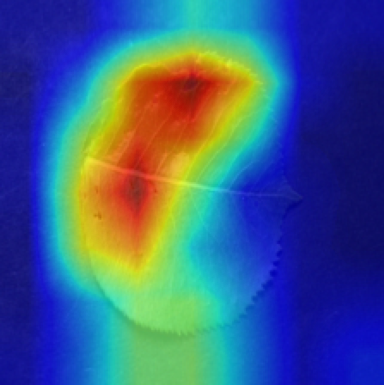} \\

     \multirow{-4}{*}{\rotatebox{90}{\footnotesize Apple Disease}} &
     \includegraphics[width=0.11\textwidth]{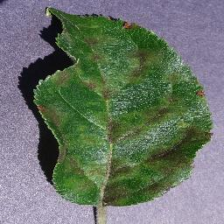} &
     \includegraphics[width=0.11\textwidth]{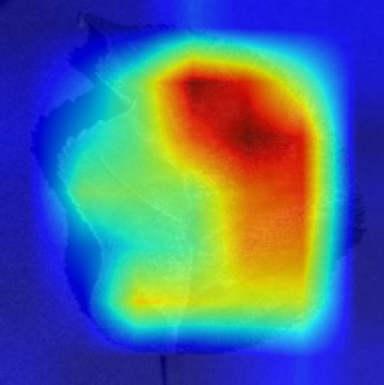} &
     \includegraphics[width=0.11\textwidth]{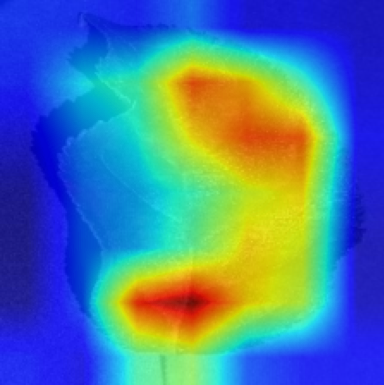} &
     \includegraphics[width=0.11\textwidth]{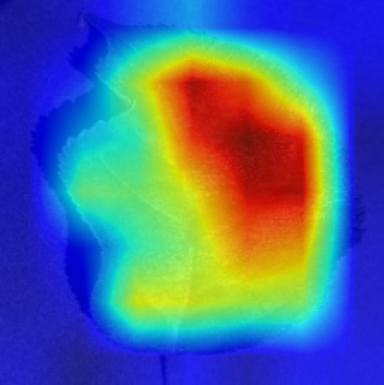} &
     \includegraphics[width=0.11\textwidth]{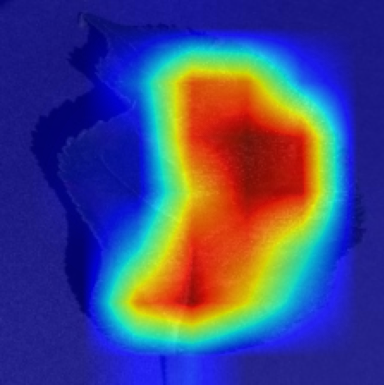} &
     \includegraphics[width=0.11\textwidth]{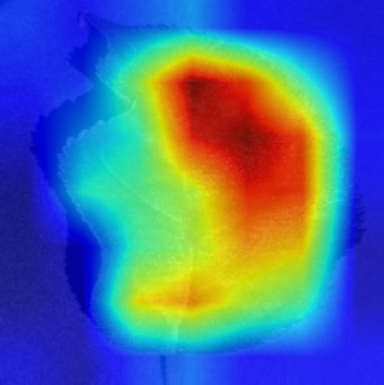} &
     \includegraphics[width=0.11\textwidth]{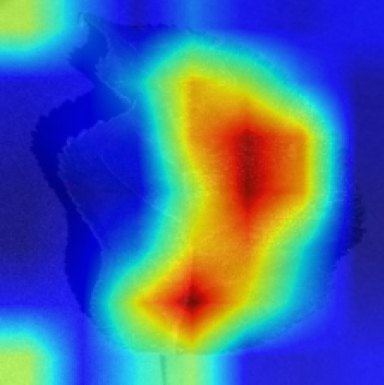} &
     \includegraphics[width=0.11\textwidth]{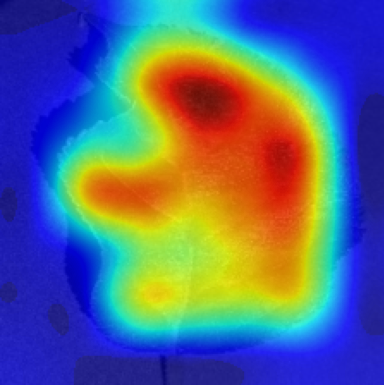} \\
    \multirow{-4}{*}{\rotatebox{90}{\footnotesize Multi-instances}} &\includegraphics[width=0.11\textwidth]{fig/ILSV2012_examples/baseball/ILSVRC2012_val_00001468_input.png} &
    \includegraphics[width=0.11\textwidth]{fig/ILSV2012_examples/baseball/ILSVRC2012_val_00001468_gradcam_pred_baseball.png} &
    \includegraphics[width=0.11\textwidth]{fig/ILSV2012_examples/baseball/ILSVRC2012_val_00001468_gradcampp_pred_baseball.png} &
    \includegraphics[width=0.11\textwidth]{fig/ILSV2012_examples/baseball/ILSVRC2012_val_00001468_xgradcam_pred_baseball.png} &
    \includegraphics[width=0.11\textwidth]{fig/ILSV2012_examples/baseball/ILSVRC2012_val_00001468_scorecam_pred_baseball.png} &
    \includegraphics[width=0.11\textwidth]{fig/ILSV2012_examples/baseball/ILSVRC2012_val_00001468_groupcam_pred_baseball.png} &
    \includegraphics[width=0.11\textwidth]{fig/ILSV2012_examples/baseball/ILSVRC2012_val_00001468_unioncam_pred_baseball.png} &
    \includegraphics[width=0.11\textwidth]{fig/ILSV2012_examples/baseball/ILSVRC2012_val_00001468_fusioncam_pred_baseball.png}

\end{tabular}
\caption{Additional qualitative visualizations comparing \textit{FusionCAM} with existing methods across different object categories and domains. \textit{FusionCAM} produces more spatially comprehensive and detailed activation maps, effectively capturing fine structural and contextual features.}
\label{fig:class_discriminative_Visualization_appendix}
\end{figure*}

Figure~\ref{fig:class_discriminative_Visualization_appendix} presents qualitative examples. The first row (ImageNet), the third row (Plant Village), and the last row (multi-instance case) are already included and discussed in the main paper. The additional rows provide further evidence of FusionCAM’s effectiveness across different datasets. In the second row (VOC2007), Fusion-CAM effectively captures both the front and side regions of the vehicle by combining the strong frontal focus of Grad-CAM with the side coverage from Score-CAM, resulting in more comprehensive localization than any single method alone.  In subsequent rows featuring different leaf species and disease symptoms, Fusion-CAM maintains fine structural details even for subtle or fragmented lesions. It produces sharper boundaries, isolates the diseased areas more distinctly, and provides a clearer visual correspondence between the pathology and the actual image features.
Overall, these supplementary visualizations confirm that FusionCAM generalizes well across categories and preserves interpretability by yielding precise and context-aware visual explanations.

\begin{table*}[h!]
\footnotesize
\centering
\begin{tabular}{|c|cc|cc|cc|cc|cc|cc|}
\hline
$\theta$ (\%) 
& \multicolumn{2}{c|}{ILSV2012} 
& \multicolumn{2}{c|}{VOC2007} 
& \multicolumn{2}{c|}{Plant Village} 
& \multicolumn{2}{c|}{Plant Leaves} 
& \multicolumn{2}{c|}{Plant K} 
& \multicolumn{2}{c|}{Apple Disease} \\
\cline{2-13}
 & AD & AI & AD & AI & AD & AI & AD & AI & AD & AI & AD & AI \\
\hline

0   & 13.73 & 41.00 & 1.72 & 28.85 & 6.24 & 12.60 & 13.26 & 4.00 & 7.96 & 7.50 & 1.11 & 8.40 \\
10  & \textbf{13.25} & \textbf{42.25} & \textbf{1.56} & \textbf{29.70} & \textbf{6.17} & 12.80 & \textbf{12.43} & 4.00 & 7.75 & 7.30 & 1.04 & \textbf{8.70} \\
20  & 13.87 & 40.70 & 1.60 & 29.20 & 6.19 & 12.70 & 13.87 & \textbf{4.40} & \textbf{7.65} & 7.30 & \textbf{1.03} & 8.60 \\
30  & 14.81 & 39.55 & 1.70 & 28.95 & 6.20 & 12.70 & 15.13 & 3.80 & 7.72 & \textbf{7.40} & 0.22 & 6.90 \\
40  & 15.27 & 38.50 & 1.69 & 29.65 & 6.25 & \textbf{13.10} & 15.24 & 3.20 & 7.67 & 7.10 & 1.14 & 7.60 \\
50  & 15.72 & 37.45 & 1.95 & 28.65 & 6.22 & 12.60 & 16.02 & 3.00 & 7.98 & 6.90 & 1.20 & 7.20 \\
60  & 16.41 & 37.10 & 2.13 & 27.30 & 6.31 & 12.80 & 16.00 & 3.00 & 8.64 & 6.80 &  1.33 & 5.80 \\
70  & 17.29 & 34.70 & 1.95 & 26.60 & 6.44 & 12.20 & 15.64 & 3.20 & 8.64 & 6.80 & 2.07 & 5.00 \\
80  & 18.28 & 33.60 & 2.06 & 26.55 &  6.58 & 11.80 & 15.31 & 3.40 & 10.66 & 6.20 & 3.97 & 4.00 \\
90  & 19.08 & 32.30 & 2.41 & 26.35 & 6.87 & 11.20  & 14.54 & 3.40 & 11.53 & 5.60 & 5.90 & 3.80 \\
\hline
\end{tabular}

\caption{Evaluation results of AD and AI metrics across the datasets for varying threshold values $\theta$. \textbf{Bold} values indicate the best performance for each dataset and metric.}
\label{tab:App_AD_AI_threshold}

\end{table*}

\section{Additional Insertion and Deletion Analysis}
\label{app:additionnal_inser_deletion}

\begin{figure*}[h!]
\centering
\setlength{\tabcolsep}{2pt} 
\renewcommand{\arraystretch}{1.2} 
\begin{tabular}{c c c c c c}
    & \footnotesize Input & \footnotesize Union-CAM & \footnotesize Fusion-CAM & \footnotesize Insertion Curves & \footnotesize Deletion Curves \\

    \multirow{-8}{*}{\rotatebox{90}{\footnotesize ILSV2012}} &
    \raisebox{0.25\height}{\includegraphics[width=0.14\textwidth]{fig/insertion_deletion/ILSVRC2012_val_00000020_input.png}} &
    \raisebox{0.25\height}{\includegraphics[width=0.14\textwidth]{fig/insertion_deletion/ILSVRC2012_val_00000020_UnionCAM.png}} &
    \raisebox{0.25\height}{\includegraphics[width=0.14\textwidth]{fig/insertion_deletion/ILSVRC2012_val_00000020_FusionCAM.png}} &
    \includegraphics[width=0.24\textwidth]{fig/inser_del/ILSVRC2012_val_00000020__insertion.png} &
    \includegraphics[width=0.24\textwidth]{fig/inser_del/ILSVRC2012_val_00000020__deletion.png} \\

    \multirow{-8}{*}{\rotatebox{90}{\footnotesize VOC2007}} &
    \raisebox{0.25\height}{\includegraphics[width=0.14\textwidth]{fig/insertion_deletion/001242_input.png}} &
    \raisebox{0.25\height}{\includegraphics[width=0.14\textwidth]{fig/insertion_deletion/001242_UnionCAM.png}} &
    \raisebox{0.25\height}{\includegraphics[width=0.14\textwidth]{fig/insertion_deletion/001242_FusionCAM.png}} &
    \includegraphics[width=0.24\textwidth]{fig/inser_del/001242_insertion.png} &
    \includegraphics[width=0.24\textwidth]{fig/inser_del/001242_deletion.png} \\

    \multirow{-8}{*}{\rotatebox{90}{\footnotesize Plant Village}} &
    \raisebox{0.25\height}{\includegraphics[width=0.14\textwidth]{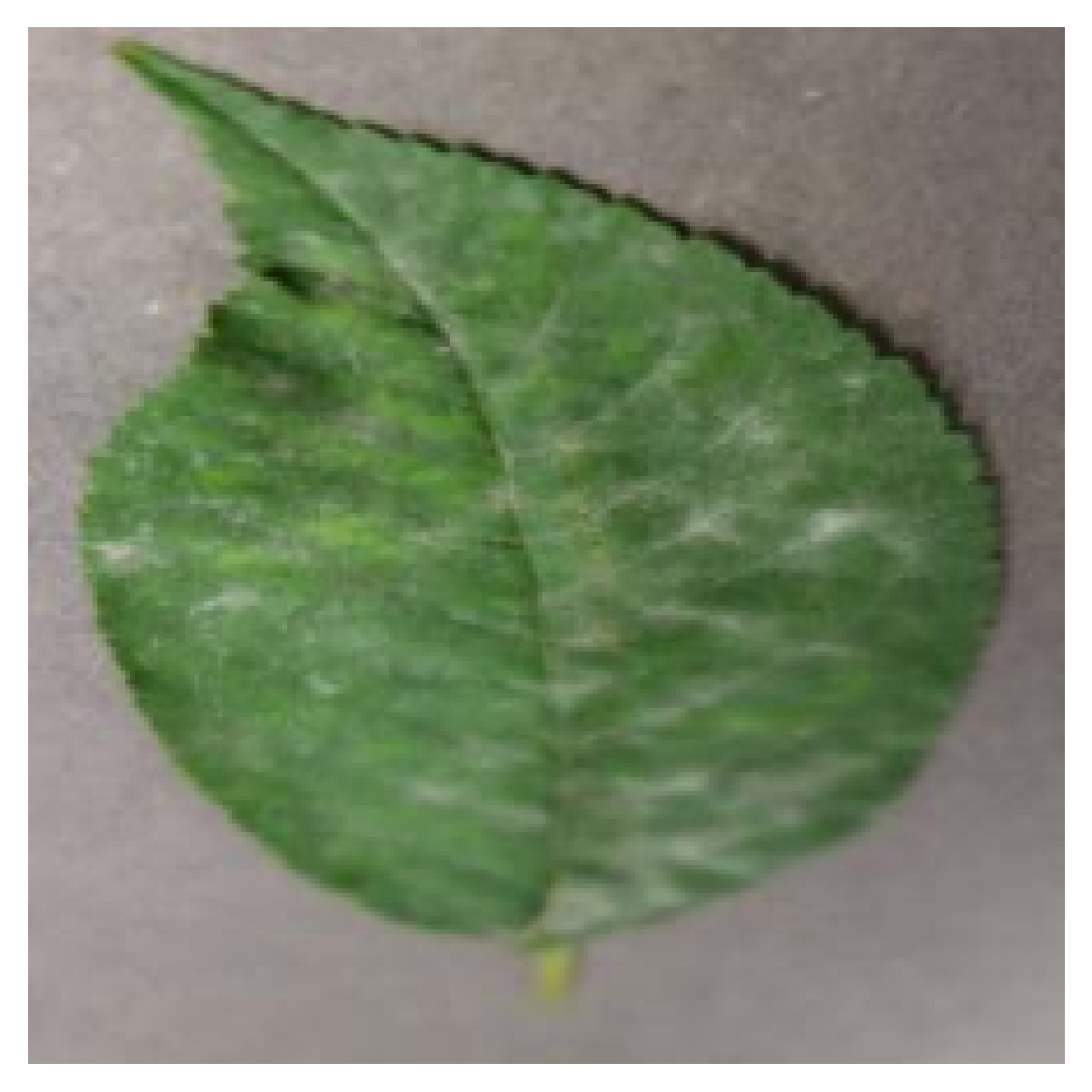}} &
    \raisebox{0.25\height}{\includegraphics[width=0.14\textwidth]{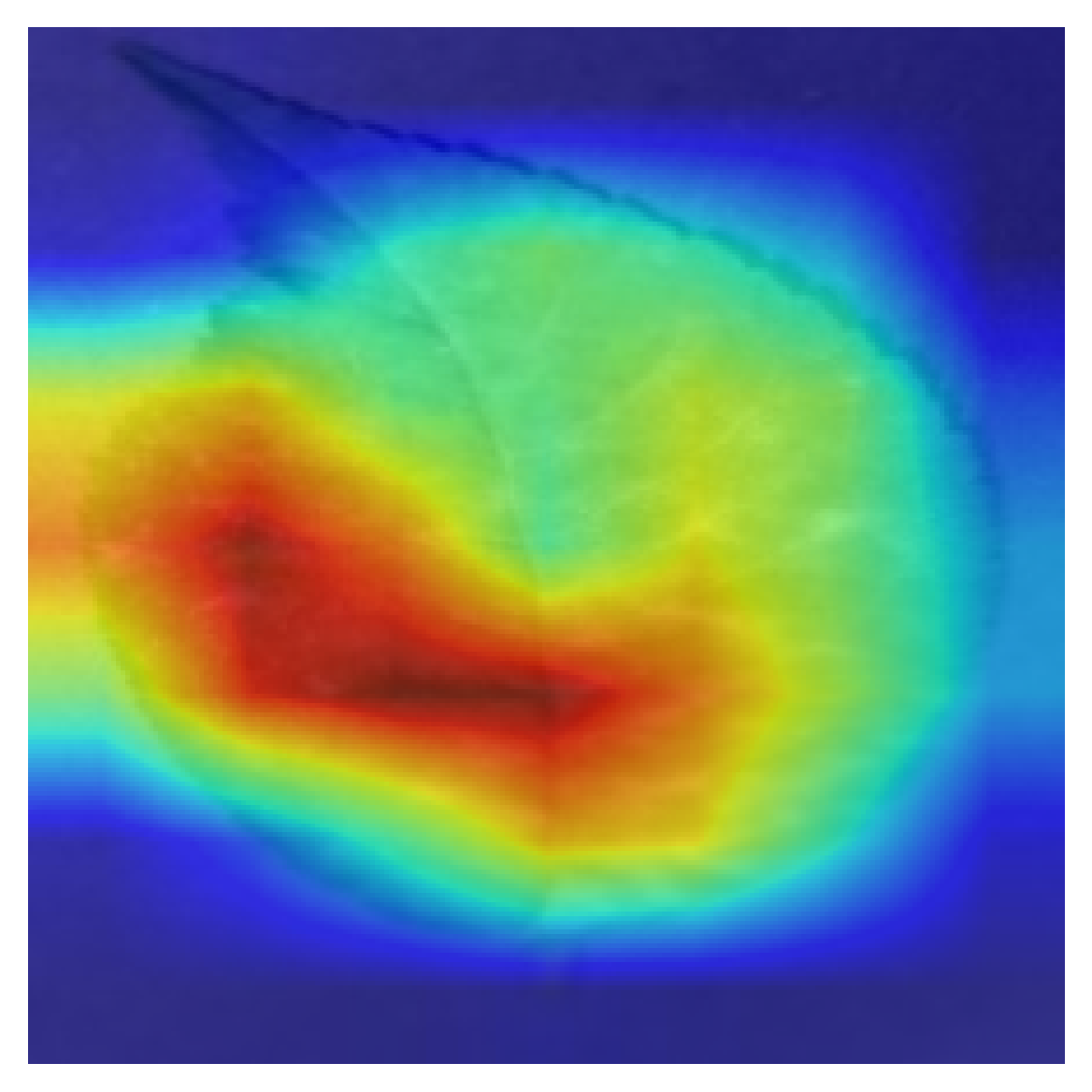}} &
    \raisebox{0.25\height}{\includegraphics[width=0.14\textwidth]{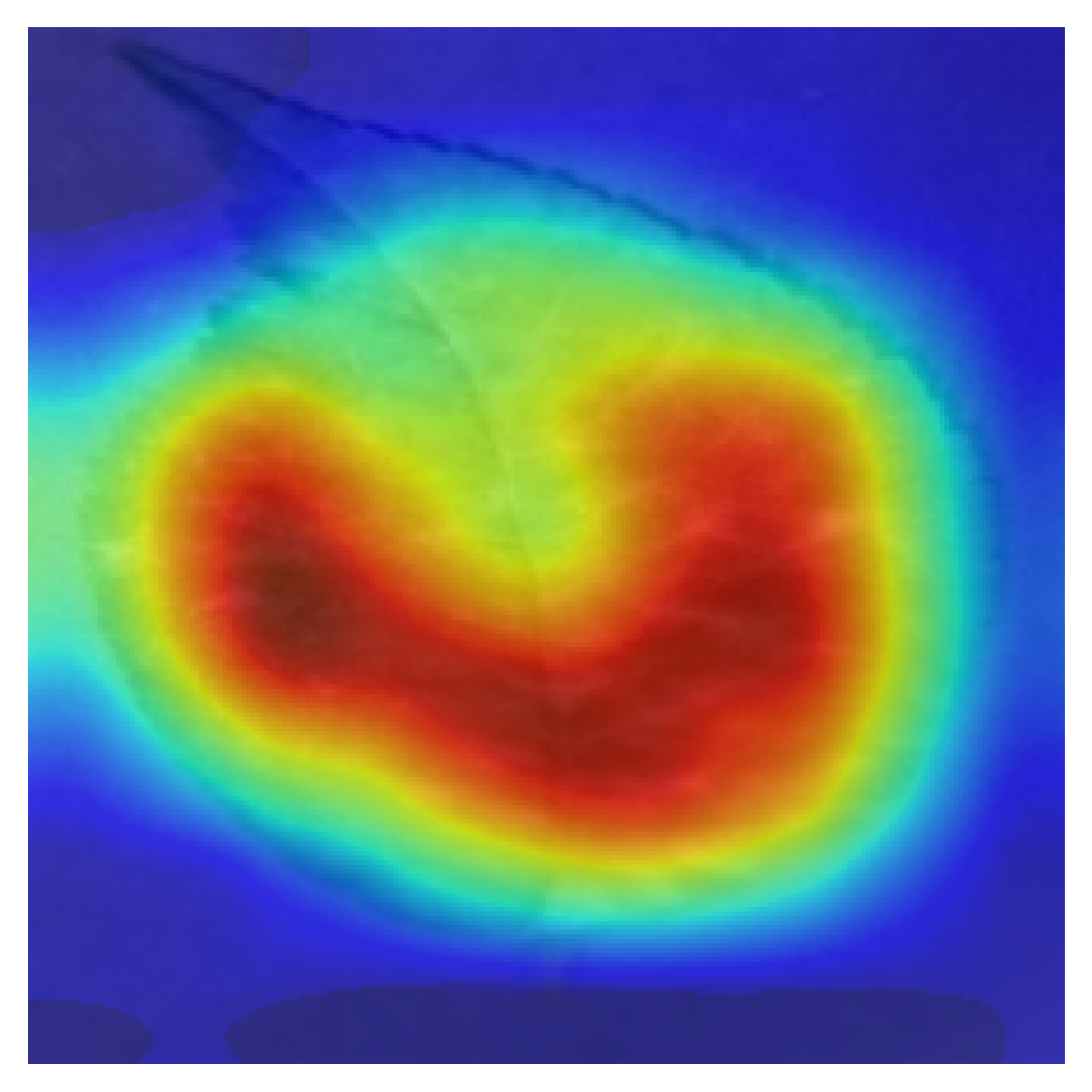}} &
    \includegraphics[width=0.24\textwidth]{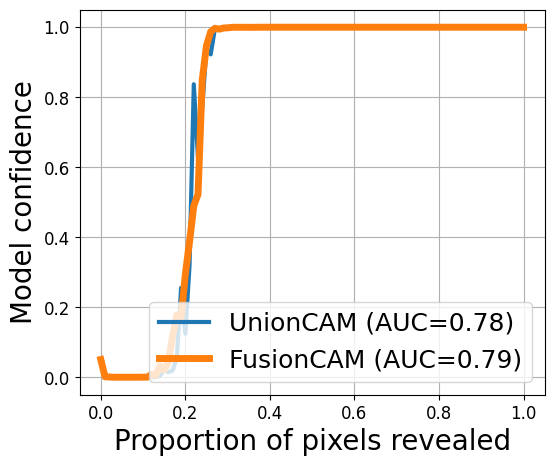} &
    \includegraphics[width=0.24\textwidth]{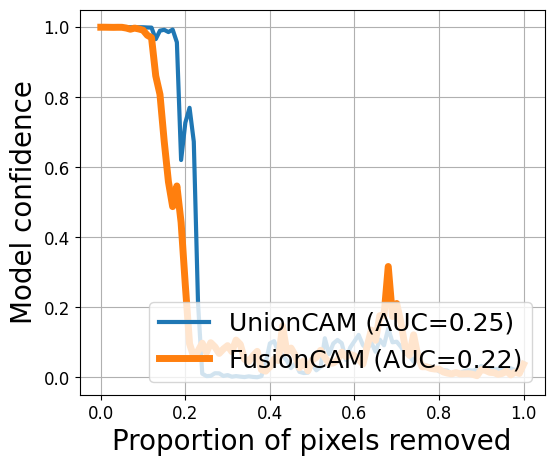} \\

    \multirow{-8}{*}{\rotatebox{90}{\footnotesize Plant Leaves}} &
    \raisebox{0.25\height}{\includegraphics[width=0.14\textwidth]{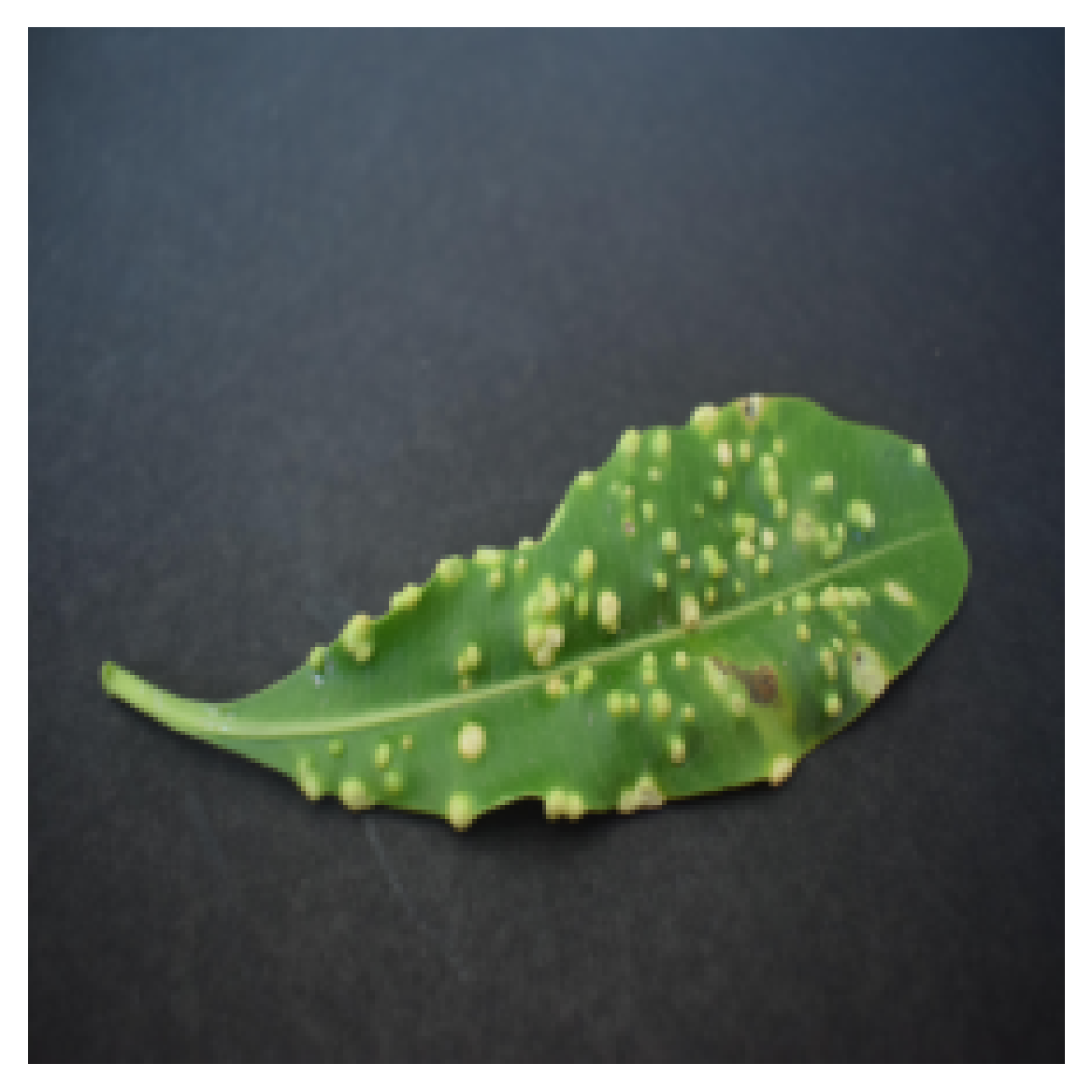}} &
    \raisebox{0.25\height}{\includegraphics[width=0.14\textwidth]{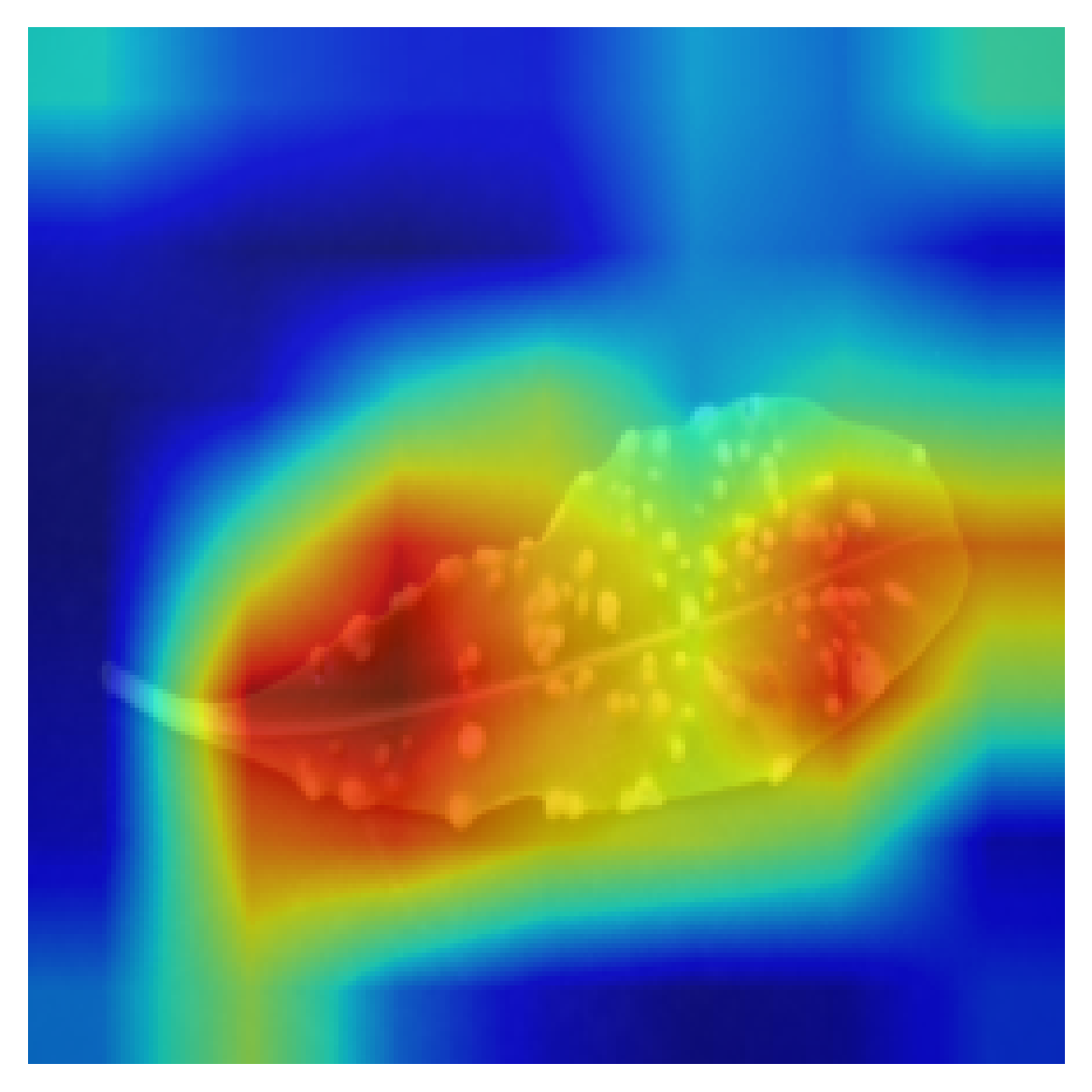}} &
    \raisebox{0.25\height}{\includegraphics[width=0.14\textwidth]{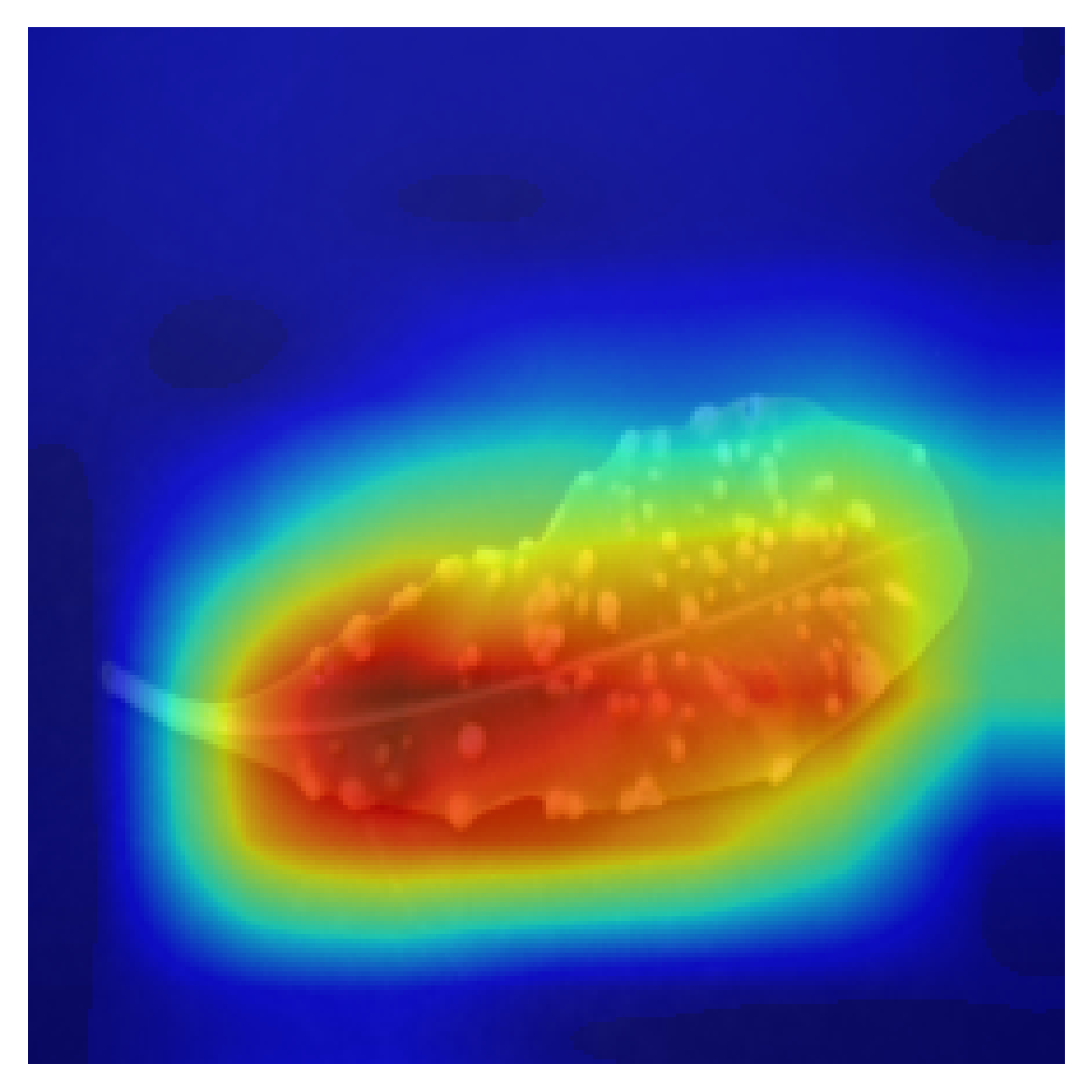}} &
    \includegraphics[width=0.24\textwidth]{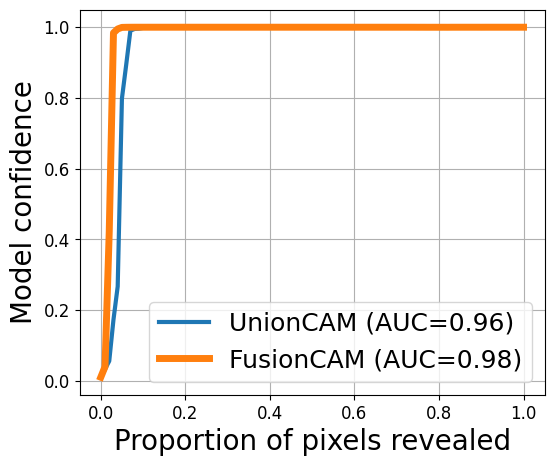} &
    \includegraphics[width=0.24\textwidth]{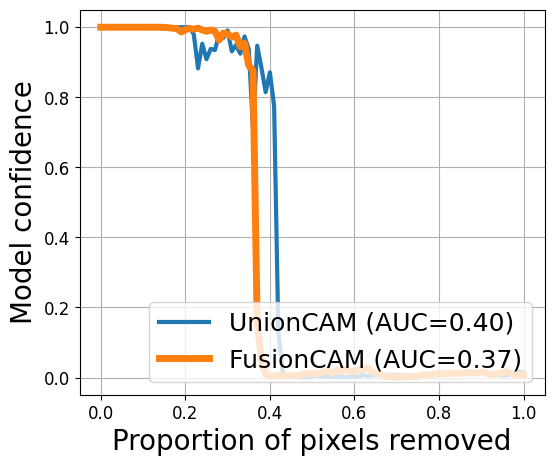} \\

    \multirow{-8}{*}{\rotatebox{90}{\footnotesize Plant K}} &
    \raisebox{0.25\height}{\includegraphics[width=0.14\textwidth]{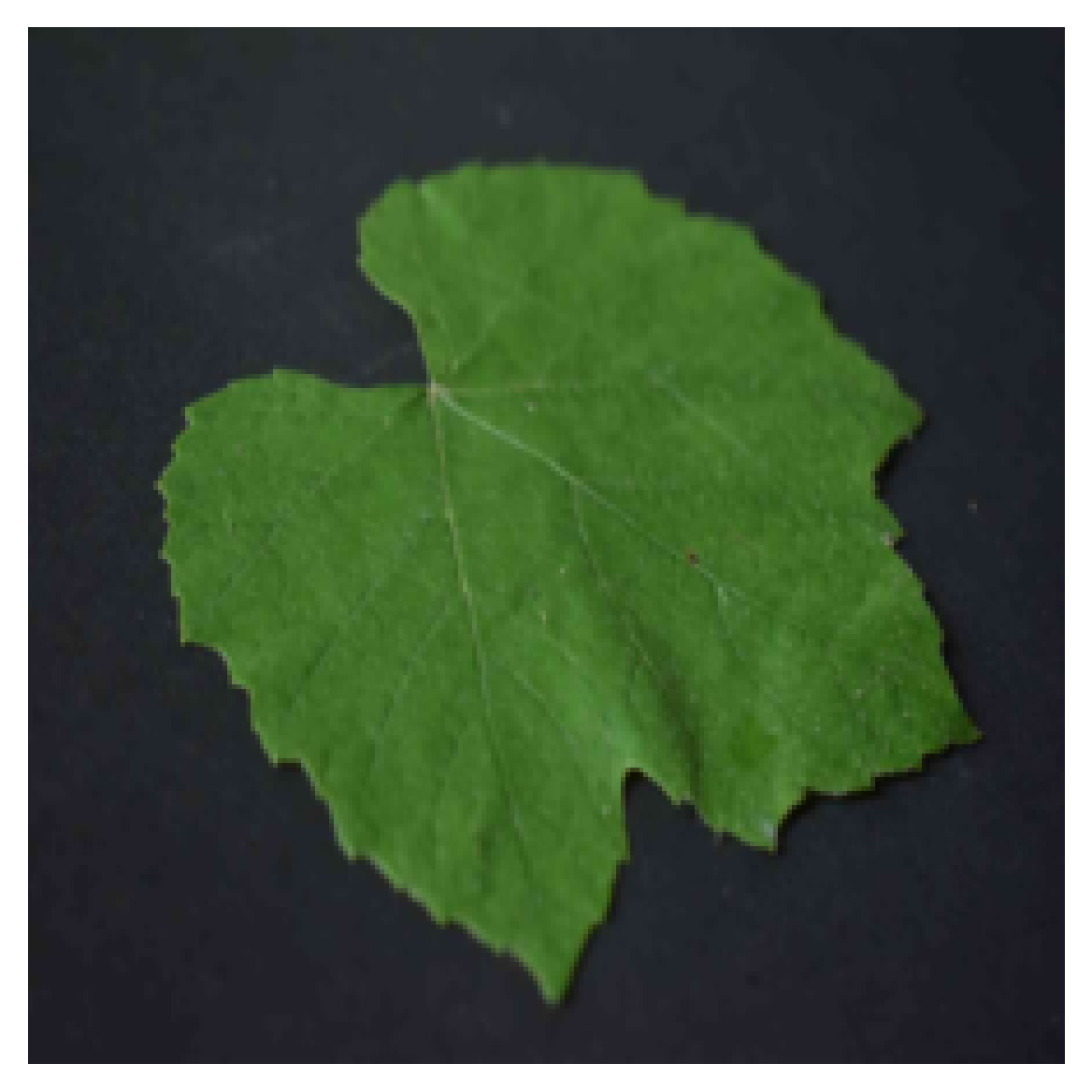}} &
    \raisebox{0.25\height}{\includegraphics[width=0.14\textwidth]{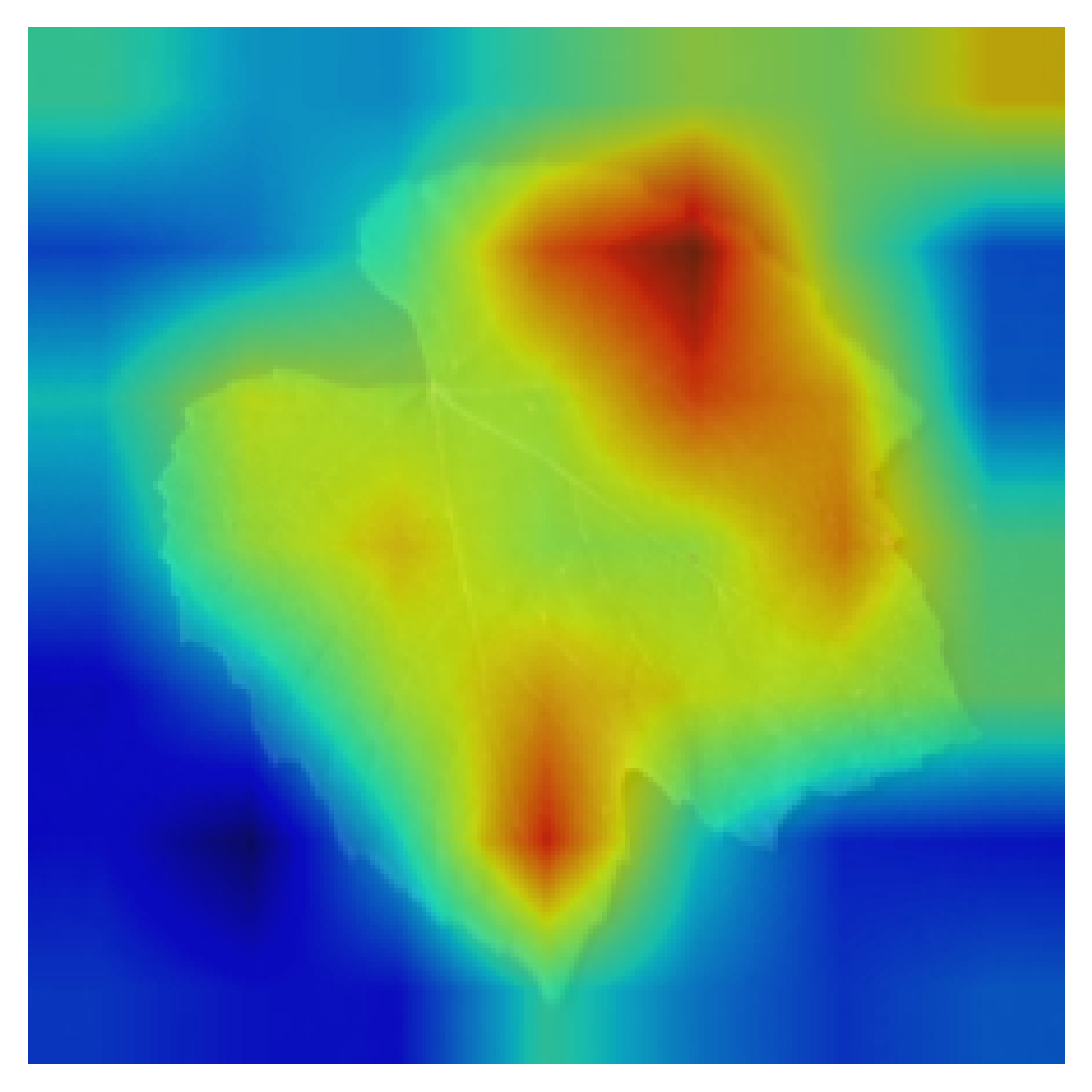}} &
    \raisebox{0.25\height}{\includegraphics[width=0.14\textwidth]{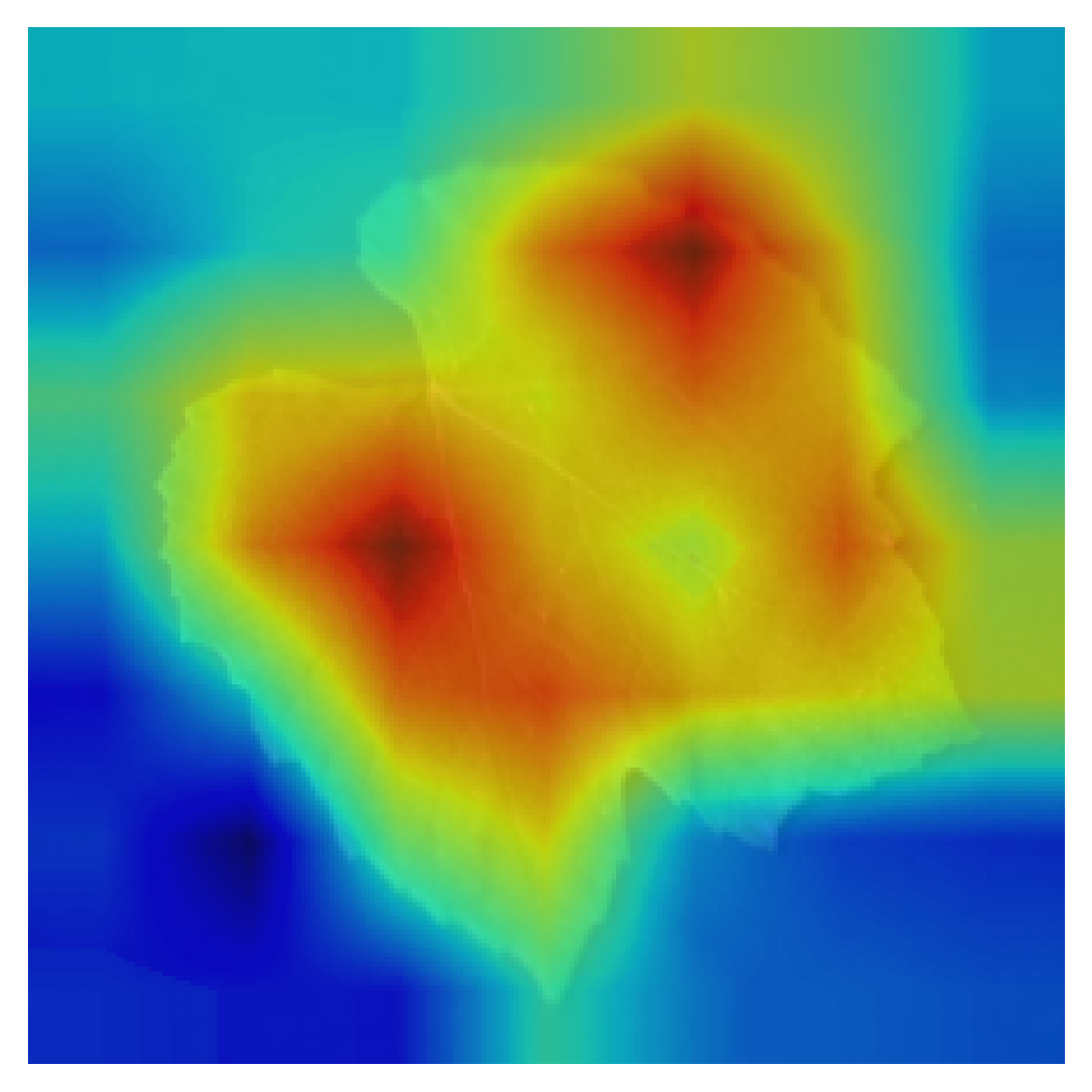}} &
    \includegraphics[width=0.24\textwidth]{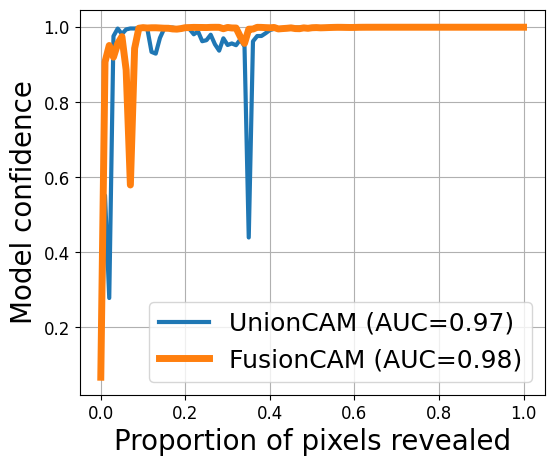} &
    \includegraphics[width=0.24\textwidth]{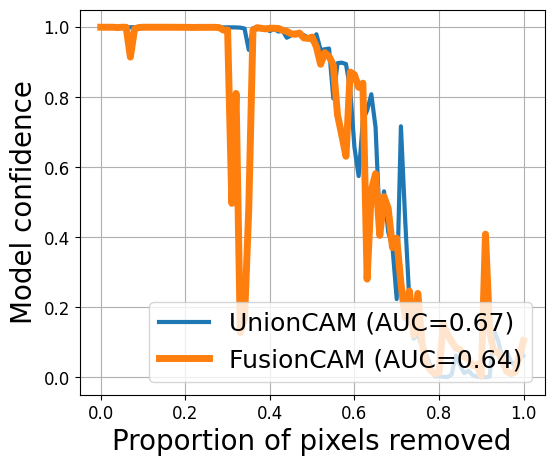} \\
\end{tabular}
\caption{Insertion and deletion curves for the ensemble CAM techniques Union-CAM and Fusion-CAM.}
\label{fig:Ins_Del_vis_appendix}
\end{figure*}

Figure~\ref{fig:Ins_Del_vis_appendix} presents additional insertion and deletion curves for various plant disease samples, complementing the results shown in the main paper.
Insertion and deletion metrics quantitatively evaluate the faithfulness of explanation maps by progressively revealing or removing the most salient pixels based on the model’s confidence score. Higher insertion AUC and lower deletion AUC indicate that the highlighted regions strongly influence the model’s prediction. Across different datasets, \textit{FusionCAM} consistently achieves competitive or superior insertion and deletion performance compared to Union-CAM, which demonstrates that FusionCAM provides both visually informative activation maps but also captures features that are causally important to the model’s decision.

\section{Threshold Parameter Analysis}
\label{app:Threshold}

Extending the threshold parameter analysis of the denoising step---which is the first step of Fusion-CAM and consists of removing the bottom $\theta\%$ of pixel intensities in the gradient-based heatmap, as low-activation values correspond to irrelevant background areas---to a threshold of 90\% in order to provide more complete understanding of the impact of the denoising threshold parameter. As represented in Table~\ref{tab:App_AD_AI_threshold}, the optimal balance between detail and focus remains at 10–20\% as discussed in the main paper. Increasing the threshold beyond this range can remove key features that contribute to the model's prediction. This explains the observed rise in Average Drop and the decrease in Average Increase, confirming the importance of selecting a lower threshold to maintain faithful and informative class activation maps.
\begin{table*}[]
\centering
\footnotesize
\begin{tabular}{|l|c|c|c|c|c|c|c|}
\hline
\textbf{XAI Method} &\textbf{ILSVRC2012} & \textbf{VOC2007}&\textbf{PlantVillage} &\textbf{Plant Leaves} & \textbf{PlantK} & \textbf{Apple Disease} & \textbf{AVG} \\ \hline

Grad-CAM     & 0.028 & 0.027 & 0.016 & 0.026 & 0.028 & 0.014 & 0.023 \\
Grad-CAM++   & 0.046 & 0.027 & 0.020 & 0.027 & 0.027 & 0.014 & 0.027 \\
\textbf{XGrad-CAM} & \textbf{0.027} & \textbf{0.015} & \textbf{0.015} & \textbf{0.024} & \textbf{0.013} & \textbf{0.014} & \textbf{0.018} \\
Score-CAM    & 4.036 & 4.085 & 1.936 & 4.022 & 4.114 & 2.077  & 3.378 \\
Group-CAM    & 0.498 & 0.554 & 0.368 & 0.610 & 0.551 & 0.293 & 0.479 \\
Union-CAM    & 5.848 & 6.140 & 2.091 & 6.435 & 6.517 & 2.339  & 4.562 \\
Fusion-CAM   & \underline{4.219} & \underline{4.199} & \underline{1.962} & \underline{4.139} & \underline{4.230} & \underline{2.118} & \underline{3.478} \\ \hline
\end{tabular}
\caption{Average generation time (in seconds) of XAI methods-using NVIDIA RTX 1080 Ti GPU. Gradient-based approaches are fastest, region-based ones are slower, and ensemble methods the most costly, with Fusion-CAM outperforming Union-CAM.}
\label{tab:app_computation_cost_xai}
\end{table*}

\section{Computational Cost}
\label{app:computational_cost}
To complement the computational analysis presented in the main paper (on ILSVRC2012, VOC2007, and PlantVillage), Table~\ref{tab:app_computation_cost_xai} reports the average generation time of the existing techniques compared to Fusion-CAM across all datasets. This comparison highlights the trade-off between interpretability and efficiency. As expected, gradient-based approaches such as XGrad-CAM are the fastest, since they require only a single backward pass to compute the saliency map. In contrast, region-based methods are slower because they rely on multiple forward passes for mask sampling. Ensemble strategies, including Union-CAM and Fusion-CAM, are computationally heavier but provide improved localization quality, with Fusion-CAM offering a strong balance between speed and accuracy.


\end{document}